\pdfoutput=1

\documentclass{article} 
\usepackage{paper,times}
\usepackage{wrapfig}
\usepackage{graphicx}
\usepackage{xcolor}
\usepackage[font=small,labelfont=bf]{caption}

\usepackage{amsmath,amsfonts,bm}

\def\eqref#1{equation~\ref{#1}}

\def\1{\bm{1}}

\DeclareMathAlphabet{\mathsfit}{\encodingdefault}{\sfdefault}{m}{sl}
\SetMathAlphabet{\mathsfit}{bold}{\encodingdefault}{\sfdefault}{bx}{n}

\usepackage{hyperref}
\hypersetup{
    colorlinks=true,
    linkcolor=red,
    filecolor=magenta,
    urlcolor=blue,
    citecolor=purple,
    pdftitle={Overleaf Example},
    pdfpagemode=FullScreen,
    }

\title{Challenges and Opportunities of Using Transformer-Based Multi-Task Learning in NLP Through ML Lifecycle: A Survey}

\author{%
\ 
\And
Lovre Torbarina\thanks{Equal contribution.}$^{\;\;,}$\thanks{Correspondence to: Lovre Torbarina \texttt{<}\href{mailto:lovre.torbarina@doxray.com}{\texttt{lovre.torbarina@doxray.com}}\texttt{>}}$^{\;\;,\;\omega}$\\
\And
Tin Ferkovic\footnotemark[1]$^{\;\;,\;\omega}$\\
\And
\ 
\AND
Lukasz Roguski$^{\;\omega}$\\
\And
Velimir Mihelcic$^{\;\omega}$\\
\And
Bruno Sarlija$^{\;\omega}$\\
\And
Zeljko Kraljevic$^{\;\omega}$\\
\AND
\hspace*{0.97\textwidth}\\
\hspace*{\fill}$^{\omega\;}$doXray B.V., Neede, Netherlands\hspace*{\fill}\\
\hspace*{\fill}\texttt{name.lastname@doxray.com}\hspace*{\fill}\\
}

\preprintcopy

\begin{document}

\maketitle

\begin{abstract}
The increasing adoption of natural language processing (NLP) models across industries has led to practitioners' need for machine learning systems to handle these models efficiently, from training to serving them in production. However, training, deploying, and updating multiple models can be complex, costly, and time-consuming, mainly when using transformer-based pre-trained language models. Multi-Task Learning (MTL) has emerged as a promising approach to improve efficiency and performance through joint training, rather than training separate models. Motivated by this, we first provide an overview of transformer-based MTL approaches in NLP. Then, we discuss the challenges and opportunities of using MTL approaches throughout typical ML lifecycle phases, specifically focusing on the challenges related to data engineering, model development, deployment, and monitoring phases. This survey focuses on transformer-based MTL architectures and, to the best of our knowledge, is novel in that it systematically analyses how transformer-based MTL in NLP fits into ML lifecycle phases. Furthermore, we motivate research on the connection between MTL and continual learning (CL), as this area remains unexplored. We believe it would be practical to have a model that can handle both MTL and CL, as this would make it easier to periodically re-train the model, update it due to distribution shifts, and add new capabilities to meet real-world requirements.
\end{abstract}

\section{Introduction}
\label{sec:introduction}
In recent years, advancements in natural language processing (NLP) have revolutionized the way we deal with complex language problems. Consequently, those advances have been significantly impacting the global industry, driving growth in organizations that incorporate AI technologies as part of their core business. To illustrate, McKinsey's state of AI report for 2022 has reported an increase of 3.8 times since 2017 in AI capabilities that organizations have embedded within at least one function or business unit, where natural language understanding (NLU) took third place among reported capabilities, just behind computer vision \citep{chui2022stateofai}. Furthermore, Fortune Business Insights projected growth of global NLP from USD 20.80 billion in 2021 to USD 161.81 billion by 2029.\footnote{\href{https://www.fortunebusinessinsights.com/industry-reports/natural-language-processing-nlp-market-101933}{Fortune Business Insights - NLP Market Size}} As a result, each NLP practitioner offering models through an API or using them internally, alone, or together with other AI capabilities, must have a machine learning (ML) system to efficiently manage these models. This involves having well-established processes from training and verifying those models to deploying them in production for end-users while continuously monitoring that those models remain up-to-date with the most recent knowledge they are being trained on.

This trend of widespread adoption of ML models by various practitioners throughout industries, and the resulting need for ML systems to manage them efficiently, was tackled in a survey of ML systems conducted by \citet{paleyes2022survey}. The survey analyzed publications and blog posts reported by different practitioners, providing insights into phases of an ML lifecycle and challenges that commonly arise during those phases. The ML lifecycle refers to the phases and processes involved in designing, developing, and deploying an ML system. It encompasses the entire process, from defining the problem and collecting data to deploying  models and monitoring their performance. Model learning and model deployment are two important phases in the ML lifecycle, among others \citep{ashmore2021assuring,paleyes2022survey}. To support practitioners' needs, the model learning phase should be equipped to handle the training and updating of a large number of models, while the model deployment phase must provide an easy and efficient way to integrate and serve those models to run in production, that is, running as part of usual business operations.

At the same time, it is a common practice in NLP production systems to utilize pre-trained language models based on transformers \citep{vaswani2017attention} by fine-tuning them for specific downstream tasks. While effective, the language models have a large number of parameters that require significant computational resources to fine-tune. Although fine-tuning pre-trained models can be more data-efficient than training a model from scratch, the expertise of annotators or domain experts may still be required to label a large number of examples, particularly if there is a significant difference between the downstream task and pre-training objectives \citep{wang2020pretrain}. Therefore, it is a costly and time-consuming procedure, especially if there is a need to train and serve multiple models in production. To address the challenge of training multiple models, researchers have been exploring Multi-Task Learning (MTL) as a solution \citep{ruder2017overview}.
MTL trains a single model to learn multiple tasks simultaneously while sharing part of the model parameters between them \citep{caruana1997multitask}, making the process memory-efficient and, in some cases, computationally more efficient than training multiple models.
Additionally, using a single model for multiple downstream tasks in a production system could simplify the integration of ML models with ML systems and reduce economic costs. This is due to the modular nature of MTL architectures that promote code and model sharing, reuse, easier collaboration, and maintenance. Moreover, the MTL model reduces idle time since the same model is used for various tasks.
Therefore, MTL approaches offer a promising solution to mitigate some of the difficulties associated with managing multiple models in ML production systems.

In this survey, we first provide an overview of transformer-based MTL approaches in NLP (see Section~\ref{section:mtl-approaches}). Second, we highlight the opportunities for using MTL approaches across multiple stages of the ML lifecycle, specifically focusing on the challenges related to data engineering, model development, deployment, and monitoring (see Section~\ref{section:mtl-and-ml-lifecyle}).
We focus solely on transformer-based architectures. To the best of our knowledge, this is the first survey that systematically discusses the benefits of using MTL approaches across multiple ML lifecycle phases (see Section~\ref{section:related-surveys}). 
Additionally, we encourage further research on the connection between MTL and Continual Learning (CL). We argue that having a model capable of handling both MTL and CL is practical as it addresses the need for periodic re-training and continual updates in response to distribution shifts and the addition of new capabilities in production models.

The rest of the paper is organized as follows. In Section~\ref{section:related-surveys}, we briefly review related surveys and highlight the gaps addressed in our survey. In Section~\ref{section:mtl-approaches}, we give an overview of transformer-based MTL approaches. In Section~\ref{section:mtl-and-ml-lifecyle}, we systematically analyze the benefits of using MTL through specific ML lifecycle phases. And finally, in Section~\ref{section:conclusion}, we give a conclusion to our work.

\section{Related Surveys}
\label{section:related-surveys}
In this section, we give an overview of related work on MTL, ML systems, and CL, and point out what has not been discussed so far regarding the connection of MTL to both ML systems and CL.

\subsection{Multi-Task Learning}
\label{subsection:mtl-surveys}

\begin{table}[t]
\caption{Discussed aspects per MTL survey. Aspects are indicated in \textbf{bold}.}
\label{table:mtl-surveys-overview}
\begin{center}
\begin{small}

    \begin{tabular}{lllllllllllllll}
        \hline
        \multicolumn{1}{c|}{Year}
        & \multicolumn{14}{c}{MTL Survey} \\
        \hline
        \multicolumn{1}{c|}{$2017$}
        & \multicolumn{14}{c}{
            $1$ - \citep{ruder2017overview}
            $2$ - \citep{zhang2017survey}
        } \\
        \multicolumn{1}{c|}{$2018$}
        & \multicolumn{14}{c}{
            $3$ - \citep{zhang2018overview}
            $4$ - \citep{thung2018review}     
        } \\
        \multicolumn{1}{c|}{$2019$}
        & \multicolumn{14}{c}{
            $5$ - \citep{zhou2019overview}  
        } \\
        \multicolumn{1}{c|}{$2020$}
        & \multicolumn{14}{c}{
            $6$ - \citep{vafaeikia2020brief}
            $7$ - \citep{worsham2020multi}
            $8$ - \citep{crawshaw2020multi}  
        } \\
        \multicolumn{1}{c|}{$2021$}
        & \multicolumn{14}{c}{
            $9$ - \citep{vandenhende2021multi}
            $10$ - \citep{chen2021multi}
            $11$ - \citep{upadhyay2021sharing}   
        } \\
        \multicolumn{1}{c|}{$2022$}
        & \multicolumn{14}{c}{
            $12$ - \citep{samant2022framework}
            $13$ - \citep{abhadiomhen2022supervised}
        } \\
        \multicolumn{1}{c|}{$2023$}
        & \multicolumn{14}{c}{
            $14$ - \citep{zhang2023survey}   
        } \\
        \hline
        \multicolumn{1}{c}{Aspect \textbackslash Survey}
        &\multicolumn{1}{c}{$1$}
        &\multicolumn{1}{c}{$2$}
        &\multicolumn{1}{c}{$3$}
        &\multicolumn{1}{c}{$4$}
        &\multicolumn{1}{c}{$5$}
        &\multicolumn{1}{c}{$6$}
        &\multicolumn{1}{c}{$7$}
        &\multicolumn{1}{c}{$8$}
        &\multicolumn{1}{c}{$9$}
        &\multicolumn{1}{c}{$10$}
        &\multicolumn{1}{c}{$11$}
        &\multicolumn{1}{c}{$12$}
        &\multicolumn{1}{c}{$13$}
        &\multicolumn{1}{c}{$14$}
        \\ \hline
        \multicolumn{1}{c}{\textbf{Computational Model}}
        &\multicolumn{14}{c}{}
        \\ \hline
        \multicolumn{1}{c|}{Traditional ML}
        &\multicolumn{1}{c}{\checkmark}
        &\multicolumn{1}{c}{\checkmark}
        &\multicolumn{1}{c}{\checkmark}
        &\multicolumn{1}{c}{\checkmark}
        &\multicolumn{1}{c}{}
        &\multicolumn{1}{c}{}
        &\multicolumn{1}{c}{}
        &\multicolumn{1}{c}{}
        &\multicolumn{1}{c}{}
        &\multicolumn{1}{c}{}
        &\multicolumn{1}{c}{}
        &\multicolumn{1}{c}{}
        &\multicolumn{1}{c}{\checkmark}
        &\multicolumn{1}{c}{}
        \\
        \multicolumn{1}{c|}{Deep Learning}
        &\multicolumn{1}{c}{\checkmark}
        &\multicolumn{1}{c}{\checkmark}
        &\multicolumn{1}{c}{\checkmark}
        &\multicolumn{1}{c}{\checkmark}
        &\multicolumn{1}{c}{\checkmark}
        &\multicolumn{1}{c}{\checkmark}
        &\multicolumn{1}{c}{\checkmark}
        &\multicolumn{1}{c}{\checkmark}
        &\multicolumn{1}{c}{\checkmark}
        &\multicolumn{1}{c}{\checkmark}
        &\multicolumn{1}{c}{\checkmark}
        &\multicolumn{1}{c}{\checkmark}
        &\multicolumn{1}{c}{}
        &\multicolumn{1}{c}{\checkmark}
        \\ \hline
        \multicolumn{1}{c}{\textbf{Architectures}}
        &\multicolumn{14}{c}{}
        \\ \hline
        \multicolumn{1}{c|}{Learning to Share}
        &\multicolumn{1}{c}{\checkmark}
        &\multicolumn{1}{c}{}
        &\multicolumn{1}{c}{}
        &\multicolumn{1}{c}{\checkmark}
        &\multicolumn{1}{c}{\checkmark}
        &\multicolumn{1}{c}{\checkmark}
        &\multicolumn{1}{c}{}
        &\multicolumn{1}{c}{\checkmark}
        &\multicolumn{1}{c}{\checkmark}
        &\multicolumn{1}{c}{\checkmark}
        &\multicolumn{1}{c}{}
        &\multicolumn{1}{c}{}
        &\multicolumn{1}{c}{}
        &\multicolumn{1}{c}{\checkmark}
        \\
        \multicolumn{1}{c|}{Universal Models}
        &\multicolumn{1}{c}{}
        &\multicolumn{1}{c}{}
        &\multicolumn{1}{c}{}
        &\multicolumn{1}{c}{}
        &\multicolumn{1}{c}{\checkmark}
        &\multicolumn{1}{c}{}
        &\multicolumn{1}{c}{}
        &\multicolumn{1}{c}{\checkmark}
        &\multicolumn{1}{c}{\checkmark}
        &\multicolumn{1}{c}{\checkmark}
        &\multicolumn{1}{c}{}
        &\multicolumn{1}{c}{}
        &\multicolumn{1}{c}{}
        &\multicolumn{1}{c}{\checkmark}
        \\ \hline
        \multicolumn{1}{c}{\textbf{Optimization}}
        &\multicolumn{14}{c}{}
        \\ \hline
        \multicolumn{1}{c|}{Loss Weighting}
        &\multicolumn{1}{c}{\checkmark}
        &\multicolumn{1}{c}{\checkmark}
        &\multicolumn{1}{c}{\checkmark}
        &\multicolumn{1}{c}{}
        &\multicolumn{1}{c}{\checkmark}
        &\multicolumn{1}{c}{\checkmark}
        &\multicolumn{1}{c}{\checkmark}
        &\multicolumn{1}{c}{\checkmark}
        &\multicolumn{1}{c}{\checkmark}
        &\multicolumn{1}{c}{\checkmark}
        &\multicolumn{1}{c}{}
        &\multicolumn{1}{c}{\checkmark}
        &\multicolumn{1}{c}{}
        &\multicolumn{1}{c}{}
        \\
        \multicolumn{1}{c|}{Regularization}
        &\multicolumn{1}{c}{\checkmark}
        &\multicolumn{1}{c}{\checkmark}
        &\multicolumn{1}{c}{\checkmark}
        &\multicolumn{1}{c}{\checkmark}
        &\multicolumn{1}{c}{\checkmark}
        &\multicolumn{1}{c}{}
        &\multicolumn{1}{c}{}
        &\multicolumn{1}{c}{\checkmark}
        &\multicolumn{1}{c}{}
        &\multicolumn{1}{c}{\checkmark}
        &\multicolumn{1}{c}{}
        &\multicolumn{1}{c}{}
        &\multicolumn{1}{c}{\checkmark}
        &\multicolumn{1}{c}{}
        \\
        \multicolumn{1}{c|}{Task Scheduling}
        &\multicolumn{1}{c}{}
        &\multicolumn{1}{c}{}
        &\multicolumn{1}{c}{}
        &\multicolumn{1}{c}{}
        &\multicolumn{1}{c}{\checkmark}
        &\multicolumn{1}{c}{\checkmark}
        &\multicolumn{1}{c}{\checkmark}
        &\multicolumn{1}{c}{\checkmark}
        &\multicolumn{1}{c}{\checkmark}
        &\multicolumn{1}{c}{\checkmark}
        &\multicolumn{1}{c}{}
        &\multicolumn{1}{c}{}
        &\multicolumn{1}{c}{}
        &\multicolumn{1}{c}{}
        \\
        \multicolumn{1}{c|}{Gradient Modulation}
        &\multicolumn{1}{c}{\checkmark}
        &\multicolumn{1}{c}{}
        &\multicolumn{1}{c}{}
        &\multicolumn{1}{c}{}
        &\multicolumn{1}{c}{}
        &\multicolumn{1}{c}{\checkmark}
        &\multicolumn{1}{c}{}
        &\multicolumn{1}{c}{\checkmark}
        &\multicolumn{1}{c}{\checkmark}
        &\multicolumn{1}{c}{\checkmark}
        &\multicolumn{1}{c}{}
        &\multicolumn{1}{c}{}
        &\multicolumn{1}{c}{}
        &\multicolumn{1}{c}{}
        \\
        \multicolumn{1}{c|}{Knowledge Distillation}
        &\multicolumn{1}{c}{}
        &\multicolumn{1}{c}{}
        &\multicolumn{1}{c}{}
        &\multicolumn{1}{c}{}
        &\multicolumn{1}{c}{}
        &\multicolumn{1}{c}{}
        &\multicolumn{1}{c}{\checkmark}
        &\multicolumn{1}{c}{\checkmark}
        &\multicolumn{1}{c}{}
        &\multicolumn{1}{c}{\checkmark}
        &\multicolumn{1}{c}{}
        &\multicolumn{1}{c}{\checkmark}
        &\multicolumn{1}{c}{}
        &\multicolumn{1}{c}{}
        \\
        \multicolumn{1}{c|}{Multi-Objective Optimization}
        &\multicolumn{1}{c}{}
        &\multicolumn{1}{c}{}
        &\multicolumn{1}{c}{}
        &\multicolumn{1}{c}{}
        &\multicolumn{1}{c}{}
        &\multicolumn{1}{c}{}
        &\multicolumn{1}{c}{}
        &\multicolumn{1}{c}{\checkmark}
        &\multicolumn{1}{c}{\checkmark}
        &\multicolumn{1}{c}{\checkmark}
        &\multicolumn{1}{c}{}
        &\multicolumn{1}{c}{}
        &\multicolumn{1}{c}{}
        &\multicolumn{1}{c}{}
        \\ \hline
        \multicolumn{1}{c}{\textbf{Task Relationship Learning}}
        &\multicolumn{14}{c}{}
        \\ \hline
        \multicolumn{1}{c|}{Task Grouping}
        &\multicolumn{1}{c}{\checkmark}
        &\multicolumn{1}{c}{\checkmark}
        &\multicolumn{1}{c}{\checkmark}
        &\multicolumn{1}{c}{\checkmark}
        &\multicolumn{1}{c}{}
        &\multicolumn{1}{c}{}
        &\multicolumn{1}{c}{\checkmark}
        &\multicolumn{1}{c}{\checkmark}
        &\multicolumn{1}{c}{\checkmark}
        &\multicolumn{1}{c}{}
        &\multicolumn{1}{c}{}
        &\multicolumn{1}{c}{}
        &\multicolumn{1}{c}{\checkmark}
        &\multicolumn{1}{c}{\checkmark}
        \\
        \multicolumn{1}{c|}{Relationships Transfer}
        &\multicolumn{1}{c}{\checkmark}
        &\multicolumn{1}{c}{}
        &\multicolumn{1}{c}{}
        &\multicolumn{1}{c}{\checkmark}
        &\multicolumn{1}{c}{}
        &\multicolumn{1}{c}{}
        &\multicolumn{1}{c}{\checkmark}
        &\multicolumn{1}{c}{\checkmark}
        &\multicolumn{1}{c}{\checkmark}
        &\multicolumn{1}{c}{}
        &\multicolumn{1}{c}{}
        &\multicolumn{1}{c}{}
        &\multicolumn{1}{c}{}
        &\multicolumn{1}{c}{}
        \\
        \multicolumn{1}{c|}{Task Embeddings}
        &\multicolumn{1}{c}{}
        &\multicolumn{1}{c}{}
        &\multicolumn{1}{c}{}
        &\multicolumn{1}{c}{}
        &\multicolumn{1}{c}{}
        &\multicolumn{1}{c}{}
        &\multicolumn{1}{c}{\checkmark}
        &\multicolumn{1}{c}{\checkmark}
        &\multicolumn{1}{c}{}
        &\multicolumn{1}{c}{}
        &\multicolumn{1}{c}{}
        &\multicolumn{1}{c}{}
        &\multicolumn{1}{c}{}
        &\multicolumn{1}{c}{}
        \\ \hline
        \multicolumn{1}{c}{\textbf{Connection to Learning Paradigm}}
        &\multicolumn{14}{c}{}
        \\ \hline
        \multicolumn{1}{c|}{Reinforcement Learning}
        &\multicolumn{1}{c}{}
        &\multicolumn{1}{c}{\checkmark}
        &\multicolumn{1}{c}{\checkmark}
        &\multicolumn{1}{c}{}
        &\multicolumn{1}{c}{\checkmark}
        &\multicolumn{1}{c}{\checkmark}
        &\multicolumn{1}{c}{}
        &\multicolumn{1}{c}{\checkmark}
        &\multicolumn{1}{c}{}
        &\multicolumn{1}{c}{}
        &\multicolumn{1}{c}{}
        &\multicolumn{1}{c}{}
        &\multicolumn{1}{c}{}
        &\multicolumn{1}{c}{\checkmark}
        \\
        \multicolumn{1}{c|}{Transfer Learning}
        &\multicolumn{1}{c}{}
        &\multicolumn{1}{c}{\checkmark}
        &\multicolumn{1}{c}{\checkmark}
        &\multicolumn{1}{c}{\checkmark}
        &\multicolumn{1}{c}{}
        &\multicolumn{1}{c}{}
        &\multicolumn{1}{c}{}
        &\multicolumn{1}{c}{}
        &\multicolumn{1}{c}{}
        &\multicolumn{1}{c}{}
        &\multicolumn{1}{c}{}
        &\multicolumn{1}{c}{}
        &\multicolumn{1}{c}{}
        &\multicolumn{1}{c}{}
        \\
        \multicolumn{1}{c|}{Meta-Learning}
        &\multicolumn{1}{c}{}
        &\multicolumn{1}{c}{}
        &\multicolumn{1}{c}{}
        &\multicolumn{1}{c}{}
        &\multicolumn{1}{c}{}
        &\multicolumn{1}{c}{}
        &\multicolumn{1}{c}{}
        &\multicolumn{1}{c}{\checkmark}
        &\multicolumn{1}{c}{}
        &\multicolumn{1}{c}{}
        &\multicolumn{1}{c}{\checkmark}
        &\multicolumn{1}{c}{}
        &\multicolumn{1}{c}{}
        &\multicolumn{1}{c}{\checkmark}
        \\
        \multicolumn{1}{c|}{Online Learning}
        &\multicolumn{1}{c}{\checkmark}
        &\multicolumn{1}{c}{\checkmark}
        &\multicolumn{1}{c}{\checkmark}
        &\multicolumn{1}{c}{}
        &\multicolumn{1}{c}{}
        &\multicolumn{1}{c}{}
        &\multicolumn{1}{c}{}
        &\multicolumn{1}{c}{}
        &\multicolumn{1}{c}{}
        &\multicolumn{1}{c}{}
        &\multicolumn{1}{c}{}
        &\multicolumn{1}{c}{}
        &\multicolumn{1}{c}{}
        &\multicolumn{1}{c}{}
        \\
        \multicolumn{1}{c|}{Continual Learning}
        &\multicolumn{1}{c}{}
        &\multicolumn{1}{c}{}
        &\multicolumn{1}{c}{}
        &\multicolumn{1}{c}{}
        &\multicolumn{1}{c}{}
        &\multicolumn{1}{c}{}
        &\multicolumn{1}{c}{}
        &\multicolumn{1}{c}{}
        &\multicolumn{1}{c}{}
        &\multicolumn{1}{c}{}
        &\multicolumn{1}{c}{}
        &\multicolumn{1}{c}{}
        &\multicolumn{1}{c}{}
        &\multicolumn{1}{c}{}
        \\ \hline
        \multicolumn{1}{c}{\textbf{Application Domain}}
        &\multicolumn{14}{c}{}
        \\ \hline
        \multicolumn{1}{c|}{Natural Language Processing}
        &\multicolumn{1}{c}{\checkmark}
        &\multicolumn{1}{c}{\checkmark}
        &\multicolumn{1}{c}{\checkmark}
        &\multicolumn{1}{c}{\checkmark}
        &\multicolumn{1}{c}{\checkmark}
        &\multicolumn{1}{c}{}
        &\multicolumn{1}{c}{\checkmark}
        &\multicolumn{1}{c}{\checkmark}
        &\multicolumn{1}{c}{}
        &\multicolumn{1}{c}{\checkmark}
        &\multicolumn{1}{c}{\checkmark}
        &\multicolumn{1}{c}{\checkmark}
        &\multicolumn{1}{c}{}
        &\multicolumn{1}{c}{\checkmark}
        \\
        \multicolumn{1}{c|}{Computer Vision}
        &\multicolumn{1}{c}{\checkmark}
        &\multicolumn{1}{c}{\checkmark}
        &\multicolumn{1}{c}{\checkmark}
        &\multicolumn{1}{c}{\checkmark}
        &\multicolumn{1}{c}{\checkmark}
        &\multicolumn{1}{c}{\checkmark}
        &\multicolumn{1}{c}{}
        &\multicolumn{1}{c}{\checkmark}
        &\multicolumn{1}{c}{\checkmark}
        &\multicolumn{1}{c}{}
        &\multicolumn{1}{c}{\checkmark}
        &\multicolumn{1}{c}{}
        &\multicolumn{1}{c}{}
        &\multicolumn{1}{c}{}
        \\ \hline
    \end{tabular}
\end{small}
\end{center}
\end{table}

The idea of MTL was explored in many research studies. In this section, we provide an overview of related MTL surveys, address various aspects of MTL, and list them along with their corresponding surveys in Table~\ref{table:mtl-surveys-overview}.
\footnote{The broader version of the table is provided in Appendix Table~\ref{table:mtl-surveys-overview-apendix}.} In the rest of the section, we just mention related surveys and go over individual MTL aspects (shown in \textbf{bold}) in more detail.\footnote{The broader version of the section is provided in Appendix~\ref{subsection:appendix-mtl-surveys}.}

Many \textbf{application domains} were studied in previous work, ranging from surveys covering multiple domains  \citep{ruder2017overview,zhang2017survey,zhang2018overview,thung2018review,vafaeikia2020brief,crawshaw2020multi,upadhyay2021sharing,abhadiomhen2022supervised}, to those dedicated to a specific domain, such as computer vision \citep{vandenhende2021multi} or natural language processing \citep{zhou2019overview,worsham2020multi,chen2021multi,samant2022framework,zhang2023survey}.
Both traditional ML and deep learning \textbf{computational models} were studied. The traditional ML was discussed primarily in older studies, while deep learning was presented in all but one study.

MTL \textbf{architectures} were widely discussed in previous work. Hard and soft parameter sharing \citep{ruder2017overview} was the most used architecture taxonomy, but recent surveys refined the taxonomies for more precise categorization \citep{crawshaw2020multi, chen2021multi}. Next, some MTL architectures were categorized as learning-to-share \citep{ruder2017sluice}, which offers a more adaptive solution by learning how to share parameters between tasks, rather than having the sharing defined a priori. Additionally, some MTL architectures were categorized as universal models, handling multiple modalities, domains, and tasks using a single model \citep{kaiser2017one, subhojeet2019omninet}.

\textbf{Optimization} techniques for MTL architectures were also widely discussed, while loss weighting was the most common approach to mitigate MTL challenges.
Techniques include loss weighting by uncertainty \citep{Kendall2018CVPR}, learning speed \citep{Liu_Liang_Gitter_2019, Zheng_2019_CVPR}, or performance \citep{Guo_2018_ECCV, Jean2019adaptive}, among others.
Next, and closely related to weighting task losses, is a task scheduling problem that involves choosing tasks to train on at each step. Many techniques were used, from simple ones that employ uniform or proportional task sampling, to the more complicated ones, such as annealed sampling \citep{stickland2019bert} or approaches based on active learning \citep{pilault2020conditionally}.
Finally, regularization approaches \citep{long2017learning,lee2018deep,Pascal_Michiardi_Bost_Huet_Zuluaga_2021}, gradient modulation \citep{lopez_paz2017gradient,sinha2018gradient}, knowledge distillation \citep{clark2019bam}, and multi-objective optimization \citep{lin2019pareto} were also applied to optimize MTL models.

\textbf{Task relationship learning} in MTL focuses on learning explicit representations of tasks or relationships between them, and typically three categories of methods were used. 
First, task grouping aims to divide a set of tasks into groups to maximize knowledge sharing during joint training \citep{standley2020which}. 
Second, transfer relationship learning determines when transferring knowledge from one task to another will be beneficial for joint learning \citep{Zamir_2018_CVPR}.
Finally, task embedding methods aim to learn task embedding space \citep{vu2020exploring}.

Previous works made \textbf{connections to other learning paradigms}, including reinforcement learning, transfer learning, meta-learning, active and online learning.
To the best of our knowledge, there had been no prior work systematically investigating connections between MTL and CL. We believe that a connection between MTL and CL represents a promising research direction, as we will motivate the need for this connection in Section~\ref{subsection:phase-model-update-problem}.

\subsection{ML Lifecycle and ML Systems}
\label{subsection:ml-lifecycle}
The unparalleled growth of advancements in ML methods in recent years, with applications in NLP, computer vision, and others, has increased the complexity of building ML systems that need to address the requirements of an ML lifecycle. Previous works reviewed the challenges of such systems, often defining ML lifecycle phases such as data management, model learning, and model deployment, to study systematically the workings of ML systems and identify challenges within and across phases
\citep{vartak2018modeldb,ashmore2021assuring,paleyes2022survey,huyen2022designing}. For example, \citet{vartak2018modeldb} defined ML lifecycle phases and analyzed model management challenges. Furthermore, \citet{ashmore2021assuring} discussed assurance of ML for each phase, while \citet{paleyes2022survey} reviewed practitioners' challenges at each phase of an ML model deployment workflow in a broader scope than previous surveys. 
In the rest of this section, we provide a few examples of challenges that occur during typical phases of the ML lifecycle.

There are many challenges that occur during different phases of the ML lifecycle. For example, data management is typically the early phase of the ML lifecycle with associated challenges such as data collection and preprocessing \citep{polyzotis2018data,sambasivan2021everyone,whang2023data}. Subsequently, model learning and verification phases take place, presenting challenges such as selecting \citep{ding2018model} and training \citep{sun2020optimization} a model, and determining the most effective method for verifying it \citep{bernardi2019150,schroder2022monitoring}, respectively. Then, a model deployment phase takes place with challenges such as model integration \citep{sculley2015hidden,renggli2019continuous} into production. Finally, the monitoring phase with challenges such as continuous model performance monitoring \citep{schroder2022monitoring} and updating a model over time \citep{ditzler2015learning,abdelkader2020towards}.

Some challenges can impact several phases of the ML lifecycle, such as collaboration among diverse teams and roles, including software and data engineers, data scientists, and other stakeholders \citep{takeuchi2020business,nahar2022collaboration,pei2022requirements,yang2022capabilities}. Furthermore, there are challenges of bias, fairness, and accountability in ethics \citep{mehrabi2021survey,kim2021machine}, various regulations set by law \citep{marchant2011growing,politou2018forgetting} and adversarial attacks in security \citep{ren2020adversarial,rosenberg2021adversarial}, among others.

Previous works addressed MTL aspects to varying degrees in specific phases, either in a straightforward manner or indirectly. However, a systematic discussion of the potential benefits of using MTL approaches to alleviate challenges across different phases of the ML lifecycle has not been conducted.

\subsection{Continual Learning}
\label{subsection:continual-learning}

CL incrementally learns a sequence of tasks, with a goal to progressively expand acquired knowledge and utilize it for subsequent learning \citep{chen2018lifelong}.
CL aims to overcome catastrophic forgetting (CF) and facilitate knowledge transfer (KT) across tasks, where CF is the degradation of performance on previous tasks when learning new ones, and KT is the ability to apply knowledge from past tasks to new tasks \citep{ke2022continual}.
Previous reviews on CL \citep{hsu2018re,de2021continual} categorized CL settings based on the marginal output and input distributions $P(Y^{(t)})$ and $P(X^{(t)})$ of a task $t$, with $P(X^{(t)}) \neq P(X^{(t+1)})$. First, class incremental learning is characterized by an expanding output space with observed class labels such that ${Y^{(t)}} \subset {Y^{(t+1)}}$ and $P(Y^{(t)}) \neq P(Y^{(t+1)})$. Second, task incremental learning (TIL), requires a task label $t$ to identify the separate output nodes $Y^{(t)}$ for the current task $t$, where ${Y^{(t)}} \neq {Y^{(t+1)}}$. Lastly, incremental domain learning defines tasks with equal class labels and probability distributions, ${Y^{(t)}} = {Y^{(t+1)}}$, and $P(Y^{(t)}) = P(Y^{(t+1)})$.

CL approaches were also categorized into three main categories based on how task-specific information is stored and utilized during the incremental learning process. First, replay methods store samples in raw format or generate pseudo-samples with a generative model, replaying them while learning a new task to mitigate forgetting and prevent previous task interference. Second, regularization-based methods, on the other hand, avoid storing raw inputs and reduce memory requirements by introducing an extra regularization term in the loss function to consolidate prior knowledge while learning new data. Third, parameter isolation methods allocate different model parameters to each task, either by adding new task-specific branches or masking out previous task parts, to prevent forgetting and maintain task-specific knowledge. We refer readers to \citet{ke2022continual} for more refined CL taxonomy and details in NLP.

In CL, MTL was usually used as an upper-bound baseline which can use all of the data from all of the tasks simultaneously \citep{de2021continual, ke2022continual}. Since CL and MTL work in different learning settings, few works tried to connect the two paradigms. \citet{sun2020ernie} presented a continual pre-training framework named ERNIE $2.0$ which incrementally builds pre-training tasks and then learns pre-trained models on these constructed tasks via continual multi-task learning. Subsequently, ERNIE $2.0$ was tested against CL and MTL pre-training approaches to evaluate the impact on abstractive text summarization task but performed similarly to other approaches \citep{kirstein2022analyzing}. In Section~\ref{subsection:phase-model-update-problem}, we motivate further research on combining CL and MTL approaches, as pre-training approaches are not sufficient for handling distribution shifts and adjusting models for new business requirements in real-world scenarios.

\section{Multi-Task Learning Approaches}
\label{section:mtl-approaches}

\subsection{Taxonomy}
There were various MTL taxonomies covered in the surveys presented in Section~\ref{subsection:mtl-surveys}.
\citet{ruder2017overview} distinguishes between a \textit{hard} and \textit{soft} parameter sharing, which proved to be an influential taxonomy, since it was used in later works as well. 
\citet{zhang2018overview} defines three categories of multi-task supervised learning -- \textit{feature-}, \textit{parameter-}, and \textit{instance-based}.
\citet{vandenhende2021multi} distinguishes between \textit{encoder-focused} and \textit{decoder-focused} architectures.
\citet{chen2021multi} discusses \textit{parallel, hierarchical, modular}, and \textit{generative adversarial architectures}.
We categorized transformer-based MTL approaches into $3$ main categories based on differences in architectures: (1) Fully-Shared Encoder, (2) Adapters, and (3) Hypernetworks (Figure~\ref{fig:mtl-architectures}).\footnote{In Appendix~\ref{subsection:appendix-prompts}, we give a brief overview of prompt engineering approaches as a fourth category. However, we do not include it in the main paper due to its need for a large number of parameters to perform well, making it inaccessible to most practitioners.}

\begin{figure}[thb]
    \centering
    \includegraphics[width=0.9\linewidth]{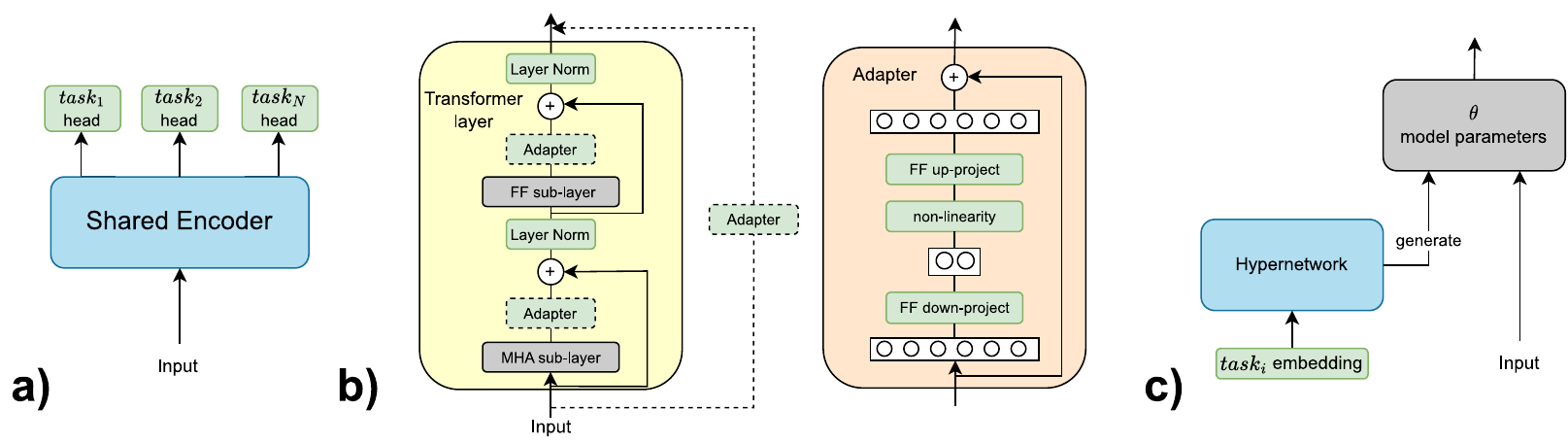}
    \caption{Simplified overview of MTL architectures. Sub-figure \textbf{a)} represents a fully-shared encoder, \textbf{b)} an adapter, and \textbf{c)} a hypernetwork. Blue components are trained jointly by all the tasks, green ones are task-specific, and grey ones are kept frozen. A dotted adapter component suggests possible adapter insertion positions.}
    \label{fig:mtl-architectures}
\end{figure}

\subsection{MTL Approaches Overview}
\label{subsection:arch-overview}

\subsubsection{Fully-Shared Encoder}
\label{subsubsection:arhc-shared-enc}

A simple and intuitive approach to MTL is to have a clear division between shared and task-specific parameters. In such an approach, there is a shared transformer-based encoder in lower layers, while the top layers consist of different, task-specific layers (heads). One such approach, MT-DNN \citep{liu2019multitask}, batches all the GLUE tasks \citep{wang2018glue} together and updates the model accordingly. The shared encoder is updated for all the instances, while the task-specific heads are updated only for the instances of a task that they are specific for.
There are several downsides to this MT-DNN approach. First, task interference is not taken into account and the authors simply hope that the tasks would interact well, although some of them are in different domains. Next, proportional random sampling is used, which could lead to underfitting on low-resource datasets. Finally, the loss is calculated in three different ways (for classification, regression, and ranking), and as a result, it has different scales. Nevertheless, all the loss functions are weighted equally. Notwithstanding these observations, their model outperforms fine-tuning a different BERT \citep{devlin-etal-2019-bert} model on most of the tasks. Additionally, they tried fine-tuning this multi-task model further on each task separately after training jointly on all the tasks, producing $N$ models for $N$ tasks. That again gave an improvement and a state-of-the-art performance at the time. However, an obvious downside is having a different model for each task.

Another shared-encoder approach is pre-finetuning. Muppet \citep{aghajanyan2021muppet} shared an encoder in an MTL on 46 diverse datasets. Heterogeneous batches have proved beneficial in handling noisy gradients from different tasks. Furthermore, to have stable training, the data-point loss was divided by log$(n)$, where $n$ denotes the cardinality of the label set for the associated task. They maintained a natural distribution of datasets as other approaches led to degraded performance. The authors found a threshold of around 15 tasks, below which downstream fine-tuning performance is degraded, and above which the performance improves linearly in the number of pre-finetuning tasks. A similar approach with added decoder, EXT5 \citep{aribandiext5}, extended the mixture to 107 supervised tasks, formatted them for encoder-decoder architectures, and performed pre-finetuning along with unsupervised T5's C4 span denoising \citep{raffel2020exploring}. Their mixture of tasks also included NLP applications such as reading comprehension, closed-book question answering, commonsense reasoning, dialogue, and summarization, among others. This showed that encoder-decoder models like T5 are capable of solving a wider range of NLP applications than encoder ones. However, task-specific models still achieve a better performance than general ones \citep{chung2022scaling}.

\subsubsection{Adapters}
\label{subsubsection:arch-adapters}

Before being used in NLP, adapter residual modules were first introduced for the visual domain \citep{rebuffi2017learning}. Adapters are small, task-specific modules that are typically inserted within network layers, but can also be injected in parallel to them. In this survey, the network is always a transformer-based architecture. Compared to transformers' sizes, they add a negligible number of parameters per task. The parameters of the original network remain frozen unless stated otherwise, resulting in a high degree of parameter sharing and a small number of trainable parameters. 
Consequently, adapters for new tasks can be easily added without retraining the transformer or other adapters. They learn task-specific layer-wise representation, are small, scalable, and shareable, have modular representations and a non-interfering composition of information \citep{pfeiffer2020adapterhub}. Since the respective adapters were trained separately, the necessity of sampling heuristics due to skewed data set sizes no longer arises \citep{pfeiffer2020adapterhub}.

\textbf{AdapterHub}. AdapterHub \citep{pfeiffer2020adapterhub} is a framework that allows dynamic usage of pre-trained adapters for different tasks and languages.\footnote{\url{https://adapterhub.ml}} The framework is built on top of the HuggingFace Transformers library and enables quick and easy adaptation of state-of-the-art pre-trained models. It allows for efficient parameter sharing between tasks by training many task- and language-specific adapters, which can be exchanged and combined post-hoc. One can choose to stack the adapters on top of each other, combine them with attention \citep{pfeiffer2020adapterfusion}, or replace them dynamically. Downloading, sharing, and training adapters require minimal changes in the training scripts. 

\textbf{Bottleneck adapters}. In AdapterHub's documentation, three different bottleneck adapter approaches were mentioned. 
Adapters can be inserted after both Multi-Head Attention (MHA) and Feed-Forward (FF) block \citep{houlsby2019parameter}, only after the FF block \citep{pfeiffer2020madx}, or in parallel to the Transformer layers \citep{he2021towards}.
Bottleneck adapters consist of a down-projection, non-linearity (typically ReLU), and an up-projection back to the original size. Residual connection is used, and layer normalization is applied afterward.

\textbf{Language adapters}. In the MAD-X framework \citep{pfeiffer2020madx}, the authors train (1) \textit{language adapters} via masked language modeling (MLM) on unlabelled target language data, and (2) \textit{task adapters} by optimizing a target task on labeled data in a source language with the most training data. Then, adapters are stacked, allowing for a zero-shot cross-lingual transfer by substituting the target language adapter at inference. Invertible adapters are introduced to tackle the mismatch between the pre-trained model's multilingual vocabulary and target language vocabulary. 
Consequently, language adapters could be useful when one had already trained a task adapter for a specific task and now needs to perform inference for the same task, but on new data from a different language.

\textbf{Other}. 
Projected Attention Layer (PAL) \citep{stickland2019bert} is a low-dimensional multi-head attention layer added in parallel to the transformer layers. 
Multi-head attention is applied on a down-projected input, after which an up-projection to an original dimension is applied. These down- and up-projection matrices are shared among layers, but not among tasks. 
The authors fine-tune a pre-trained encoder along the PALs. 
This has downsides: (1) forgetting of pre-trained knowledge is possible, (2) access to all the tasks at training time is required, and (3) adding new tasks requires complete joint retraining. 
Thus, this approach misses many of the characteristics of an adapter.

AdapterFusion \citep{pfeiffer2020adapterfusion} introduces a knowledge composition phase, in which the previously trained adapters are combined. This approach uses multiple adapters to maximize knowledge transfer between tasks without suffering from the MTL drawbacks, such as catastrophic forgetting \citep{serra2018overcoming} or task interference \citep{wu2020understanding}. It introduces a new set of weights that learn to combine the adapters as a dynamic function of the target task data by using attention. This shows its greatest downside -- AdapterFusion is trained for one task only. 

\citet{hu2021lora} argue that the original adapter bottleneck design \citep{houlsby2019parameter} introduces inference latency because the adapters are processed sequentially, whereas large language models (LLMs) rely on hardware parallelism. 
Their approach, LoRA (Low Rank Approximation) modifies attention weights of query and value projection matrices by introducing trainable low-rank decomposition matrices in parallel to the original computation.
This reduces inference latency, as the decomposition matrices can be merged with the pre-trained weights for faster inference. 

\begin{minipage}{\textwidth}
    \begin{minipage}[b]{0.44\textwidth}
        \captionsetup{type=figure}
        \begin{center}
        \includegraphics[width=0.7\textwidth]{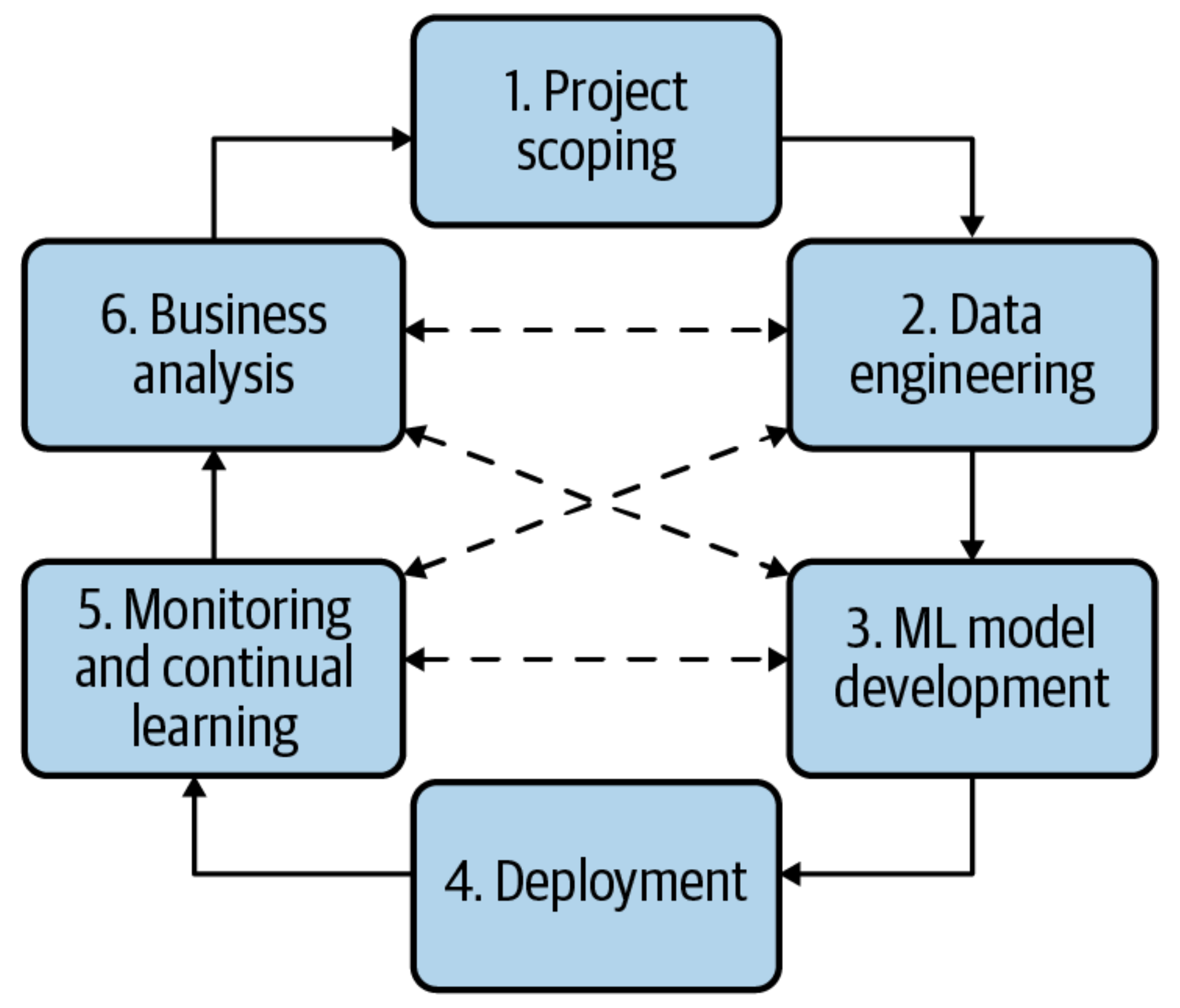}
        \end{center}
        \caption{ML lifecycle phases. Image is taken from \\ \citet{huyen2022designing}.}
        \label{fig:ml-lifecycle-phases}
    \end{minipage}
    \hfill
    \begin{minipage}[b]{0.55\textwidth}
        \captionsetup{type=table}
        \caption{ML Lifecycle phases and corresponding challenges.}
        \label{table:ml-lifecycle-phase-challenges}
        \begin{center}
        \begin{small}
        \begin{tabular}[b]{l|l}
            \hline
            \textbf{ML Lifecycle Phase} & \textbf{Challenges} \\ \hline
            Project Scoping & Initial Requirements \\ \hline
            Data Engineering & Labeling of Large Volumes of Data \\
            & Cost of Annotators and Experts \\
            & Lack of High-Variance Data \\ \hline
            Model Development & Model Complexity \\
            & Resource-Constrained Environments \\
            & Computational Cost \\
            & Environmental Impact \\ \hline
            Deployment & Ease of integration \\ \hline
            Monitoring & Distribution Shift \\ \hline
            Business Analysis & New Requirements \\ \hline
        \end{tabular}
        \end{small}
        \end{center}
    \end{minipage}
\end{minipage}

\subsubsection{Hypernetworks}
\label{subsubsection:arch-hypernet}

A hypernetwork is a network that generates weights of another network \citep{ha2016hypernetworks}. 
This approach can alleviate a downside of adapters, which is a lack of knowledge sharing. 
Hypernetwork allows sharing of knowledge across tasks while adapting to individual tasks through task-specific parameter generation.

CA-MTL \citep{pilault2020conditionally} modularizes a pre-trained network by either adding task-conditioned layers or changing the pre-trained weights using the task embedding. Their task-conditioned transformer-based network has four components: (1) conditional attention, (2) conditional alignment, (3) conditional layer normalization, and (4) conditional bottleneck. In (1), they use block-diagonal conditional attention which allows the attention to account for task-specific biases. Component (2) aligns the data of diverse tasks. In (3), they adjust layer normalization statistics for specific tasks. Finally, (4) facilitates weight sharing and boosts task-specific information flow from lower layers. 
In addition, they use multi-task uncertainty sampling. This favors tasks with the highest uncertainty by sampling a task whenever its entropy increases, helping to avoid catastrophic forgetting. When introducing a new task, they claim that only a new linear decoder head and a new task embedding vector need to be added to re-modulate existing weights.

HyperFormer++ \citep{mahabadi2021parameter} uses hypernetworks to generate the weights of adapter and layer normalization parameters. These hypernetworks condition on a task embedding, adapter position (after MHA or FF sub-layer), and layer id within the T5 model \citep{raffel2020exploring}. During training, they sample the tasks using temperature-based sampling. They state that for each new task, their model only requires learning an additional task embedding. 

HyperGrid \citep{tay2020hypergrid} leverages a grid-wise decomposable hyper projection structure which helps specialize regions in weight matrices for different tasks. To construct the proposed hypernetwork, their method learns the interactions and composition between a global, task-agnostic state and a local, task-specific state. They equip position-wise FF sub-layers of a Transformer with HyperGrid. They initialize a T5 model from a pre-trained checkpoint and add additional parameters that are fine-tuned along the rest of the network. The authors of the paper did not mention anything specific regarding the ability to add a new task without re-training, as it appears to be non-trivial.

\section{MTL From ML Lifecycle Point of View}
\label{section:mtl-and-ml-lifecyle}

In this section, we discuss the challenges and opportunities of incorporating into ML production systems MTL approaches instead of using multiple single-task counterparts. Motivated by previous reviews on ML lifecycle (see Section~\ref{subsection:ml-lifecycle}), we define ML lifecycle phases to discuss challenges and opportunities in a systematic manner. 

Following \citet{huyen2022designing}, we define six ML lifecycle phases: (1) Project Scoping, (2) Data Engineering, (3) Model Development, (4) Deployment, (5) Monitoring, and (6) Bussiness Analysis (Figure~\ref{fig:ml-lifecycle-phases}). Next, we mostly focus on challenges that were discussed in a broader scope in \citet{paleyes2022survey}, while we argue how MTL can alleviate them. Challenges per each phase of the ML lifecycle are listed in Table~\ref{table:ml-lifecycle-phase-challenges}. When discussing the challenges and opportunities of using MTL approaches, we compare them to corresponding single-task model solutions.

In the rest of the section, we first discuss data engineering and model development phases in isolation. Then, we discuss an ML model updating problem by indicating how certain aspects of the problem pose different challenges in different phases of the ML lifecycle.

\subsection{Data Engineering}
\label{subsection:phase-data-engineering}
The first phase we discuss is data engineering. This phase focuses on preparing data that is needed to train a machine learning model, while we have a particular interest in challenges related to the lack of labeled data (see Table~\ref{table:ml-lifecycle-phase-challenges}).

\textbf{Challenges.} The need for data augmentation can arise from various factors, with one of the most problematic being the lack of labels in the data, especially in real-world applications where labeled data may be scarce. It is a common practice in NLP production systems to utilize pre-trained transformer-based language models by fine-tuning them for specific downstream tasks. However, if there is a significant gap between a downstream task and the pre-training objectives, a larger amount of labeled data may still be required to achieve the target performance \citep{wang2020pretrain}. Obtaining this data involves costly and time-consuming involvement of annotators and domain experts. Additionally, the absence of high-variance data results in a model that is unable to generalize well, such as adapting language models to low-resource languages \citep{clark2019dont}.

\textbf{Opportunities.} The challenges posed by the lack of labeled data can be alleviated using MTL approaches. For example, if a set of single-task models is being used in a production system, training an MTL model instead can help alleviate data sparsity by jointly learning to solve related tasks (Section~\ref{subsubsection:arhc-shared-enc}). The benefits of MTL have been previously discussed in \citet{caruana1997multitask, ruder2017overview}, including its ability to increase data-efficiency. First, different tasks transfer different knowledge aspects to each other, enhancing the representation's ability to express the input text, which can be beneficial for tasks with low-resource datasets (Section~\ref{subsubsection:arhc-shared-enc}).
However, some MTL approaches may underperform in resource-constrained environments due to inadequate optimization choices (Section~\ref{subsubsection:arhc-shared-enc}). 
Additionally, the presence of different noise patterns in each task acts as an implicit data augmentation method, effectively increasing the sample size used for training, and leading to a robust model with more general representations \citep{ruder2017overview}. 
Finally, using pre-finetuning (Section \ref{subsubsection:arhc-shared-enc}) could reduce convergence time, saving computational resources.

\subsection{Model Development}
\label{subsection:phase-model-development}
In the model development phase, we focus on two groups of challenges. The first group refers to the \emph{model selection} problem, including issues related to model complexity and resource constraints. The second group of challenges is related to problems in \emph{model training}, such as the computational cost of the training procedure and its impact on the environment. We refer readers to \citep{gupta2022compression} for an overview of methods for efficient models in text.

\textbf{Model Selection Challenges.} When selecting a model to handle tasks that end-users are interested in, practitioners often face a dilemma regarding the trade-off between model complexity and performance. Typically, complex models have better performance, but they come with the risk of over-complicating the design in the first place, leading to a longer development time and deployment failure \citep{haldar2019applying}. Furthermore, they may not be practical to use in resource-constrained environments where they require high computational and memory resources.

\textbf{Model Selection Opportunities.} MTL architectures, presented in Section~\ref{subsection:arch-overview}, have properties that can alleviate challenges related to the trade-off between model complexity and performance. For example, consider replacing N single-task models with a single MTL shared encoder model. The MTL model will have close to N times smaller memory footprint, as the number of task-specific parameters is negligible compared to the number of shared parameters. This reduction results in a better fit to memory-constrained environments while only having slightly worse performance than the single-task counterparts. Similarly, saving $N$ adapters or a single hypernetwork is much more memory-efficient than saving N single-task models.

\textbf{Model Training Challenges.} The training of machine learning models presents several challenges that must be addressed by practitioners. 
 One of the major challenges is the high economic cost associated with training, which is due to the computational resources required. In the field of natural language processing, the cost of model training continues to rise, even as the cost of individual floating-point operations decreases, due to factors such as the growth in training dataset size, number of model parameters, and number of operations involved in the training process \citep{sharir2020cost}. The training process also has a significant impact on the environment, leading to increased energy consumption and greenhouse gas emissions \citep{strubell2020energy}. These challenges emphasize the need to address the economic and environmental implications of training machine learning models.

\textbf{Model Training Opportunities.} Computational resource challenges could be alleviated in some aspects by using MTL approaches to reduce the cost of model training. First, in certain cases, smaller dataset sizes can be used due to pre-finetuning or knowledge transfer between related tasks during joint training of the MTL model, leading to more data-efficient training. Second, the joint model is more parameter-efficient, resulting in a significant reduction in the number of parameters required for multiple single-task models.

\textbf{Selection, Training, and Inference Trade-Off.} The number of floating-point operations is not reduced in some cases, and depends on the choice of MTL architecture and the nature of the tasks. 
Different task-specific heads require different inputs if tasks belong to different domains or have different input encodings. However, if tasks belong to the same domain and have the same input encoding, part of the computation can be shared among the task-specific heads.
For example, in a fully-shared encoder (Section~\ref{subsubsection:arhc-shared-enc}), the computation of the full encoder can be shared, while the computation in the task-specific heads is negligible.
Similarly goes for adapters (Section~\ref{subsubsection:arch-adapters}) -- the large majority of the computation is shared, and only the adapters are then dynamically plugged for performing different tasks on the same input. 
Namely, the benefit of using adapters is a small number of trainable parameters, thanks to a frozen encoder, which results in a faster gradient back-propagation. A forward pass, on the other hand, takes more time compared to an encoder with no adapters. Thus, when tasks don't share the datasets, using adapters results in a longer inference time compared to single-task counterparts. AdapterFusion inference is even slower, as an input must pass through all the available adapters. LoRA solves the inference latency problem by using parameter composition.

\subsection{Model Deployment}
\label{subsection:phase-model-deployment}
In the model deployment phase, our focus is on simplifying the integration of trained ML models with existing ML systems that are running in production, with a particular emphasis on straightforward implementation, seamless collaboration, and ease of maintenance.

\textbf{Challenges.} The first challenge in model deployment is preparing the developed model for use in a production environment. The initial versions of models are often developed by stakeholders (e.g., ML researchers) who are different from those responsible for deploying them into production (e.g., ML and DevOps engineers). This means that the model code needs to be adapted to meet the requirements of the operational production environment.
Those requirements are typically more strict and different from those available during the development phase, so it is important to address operational aspects such as scalability, security, and reliability. Hence, each additional model adds complexity to the process, both for the stakeholders involved and for the ML system infrastructure and computational resources in place.
Another challenge during model integration is incorporating the model into real-data processing pipelines in production, whether it is for batch offline processes or handling real-time user requests. During model training, the researcher often uses preprocessed and clean datasets. However, when integrating the model into production, it will be integrated into existing data pipelines. The more complex the model's input data requirements, or the more models that are used, the more complex the data processing pipelines will need to be.
Adapting the models and existing data pipelines often requires collaboration between different stakeholders or teams responsible for different parts of the ML system and/or ML lifecycle phase. All of these factors directly impact the increasing costs of maintenance, operations, support, and infrastructure.

\textbf{Opportunities.} MTL approach was presented by the team from Pinterest, where the authors trained a universal set of image embeddings for three different models, which simplified their deployment pipelines and improved performance on individual tasks \citep{zhai2019learning}.
The key benefit is that the use of MTL can reduce the number of parameters needed to support different models required to solve a problem. This leads to smaller task-specific models and less code to be adapted to the production environment, reducing the overhead of potential changes in existing data pipelines. Reusing data, code, and models can save time and simplify the model deployment process. 
Next, the modular design of MTL architectures makes it easier to work with by enabling code reuse.
For example, the fully-shared encoder architecture (described in Section~\ref{subsubsection:arhc-shared-enc}) reuses the encoder between multiple task-specific heads. 
Moreover, freezing the shared encoder would allow for working on separate tasks independently and simultaneously, making cross-team collaboration easier and more efficient. 
This idea was used in the HydraNet MTL model by the Tesla AI team \citep{tesla-2021-ai-day}. 
However, freezing an encoder and updating only the task-specific heads may decrease performance. 
Additionally, the concept behind the AdapterHub (described in Section~\ref{subsubsection:arch-adapters}) was designed to allow users to choose from a set of adapter modules, combine them in their preferred way, and insert or replace them dynamically into state-of-the-art pre-trained models.
To conclude, the modular nature of MTL architectures has a positive impact on making the incorporation of these architectures into ML software easier, making it simpler to develop, collaborate, configure, and integrate into deployment processes.

\subsection{Model Updating Through Multiple ML Lifecycle Phases}
\label{subsection:phase-model-update-problem}
Often it is necessary to update the ML model regularly after it has been deployed and is running in production, in order to keep it aligned with the most current changes in data and environment. The need for updating models is one of the most important requirements of ML production systems \citep{pacheco2018towards,abdelkader2020towards,lakshmanan2020machine,paleyes2022survey,huyen2022designing,wu2022survey,nahar2022collaboration}. In this section, we discuss the challenges of model updating and how MTL approaches can alleviate these challenges. Challenges of different phases of the ML lifecycle can trigger the need for model updates. First, we identify these occurrences and discuss the actions they trigger. Then, we discuss how MTL approaches can alleviate these challenges and point out the current limitations of these approaches.

\textbf{Model Updating Challenges.} A distribution shift is one of the common reasons for the need to update models. Distribution shift refers to changes observed in the joint distribution of the input and output variables of an ML model \citep{ditzler2015learning}.
Two problems must be solved to address the distribution shift effectively. First, in the \emph{monitoring phase}, a mechanism must be in place to detect changes in distribution or a drop in key performance indicators, which will signal the need for an ML model update. Second, in the \emph{model development phase}, there must be a way to continuously learn and update the model in response to signals from the monitoring phase.

New business requirements often arise, requiring the model to have new capabilities in addition to the existing ones. For example, a named entity recognition model trained for 10 entity labels in news articles may require 5 new labels after 3 months of use.
New requirements are introduced during the \emph{business analysis phase}, which also triggers the need for an ML model update. Unlike the initial requirements in the \emph{project scoping phase}, adding new requirements should be possible without re-training the full model.

Periodic or scheduled re-training and continual learning are the most common approaches for adapting models to new data and requirements. Periodic re-training refers to the process of re-training a model at a predetermined interval, regardless of whether there have been any changes to the data or the environment. The frequency of updating is determined beforehand and is typically based on the amount of data and the desired level of model performance. Striking a balance between updating the model regularly to maintain good performance and avoiding over-updating the model to minimize computational costs is crucial. It is important to note that different models may require different re-training schedules, making it a situation-dependent problem. Finding the optimal re-training schedule requires careful consideration of the specific requirements and circumstances of each ML system. Continual learning, on the other hand, refers to the ability of an ML model to adapt to changes in the data and environment over time. This approach involves continuously monitoring the performance of the model and updating it as needed to ensure it remains accurate and up-to-date. The frequency of model updates in continual learning is determined dynamically based on the changes observed in the data and environment.

\textbf{Model Updating Opportunities.} Following \citet{huyen2022designing}, we differentiate two types of model updates: \emph{data iteration} and \emph{model iteration}. Data iteration refers to updating the model with new data while keeping the model architecture and features the same, while model iteration refers to adding new features to an existing model architecture or changing the model architecture itself. To discuss the challenges and opportunities of MTL approaches for model updating, we consider two scenarios. First, periodic re-training is done every 6 or 12 months, with the goal of training each model to perform optimally using all available data. Due to the high economic cost, we assume this cannot be done more frequently. Second, between periodic retrains, there may be situations where data iteration is necessary due to a distribution shift, or model iteration is required to extend model capabilities in response to a new business requirement.

We believe that incorporating MTL approaches into the update scenario would be practical, in addition to the benefits discussed in  Sections~\ref{subsection:phase-data-engineering}-\ref{subsection:phase-model-deployment}. MTL is particularly well-suited for periodic re-training scenarios where the objective is to obtain a single best-performing model on all tasks using all available data, instead of training individual models for each task. However, the challenge lies in how MTL approaches can manage the second scenario where the MTL model must learn new tasks or domains sequentially over time. As a result, the same model must be able to learn in both MTL and CL settings as needed, and we refer to this setting as Continual MTL (CMTL).

\textbf{Continual MTL.} MTL and CL models both learn multiple tasks, however, MTL learns them simultaneously, while CL learns them incrementally. As mentioned in Section~\ref{subsection:continual-learning}, the authors of \citet{sun2020ernie} combined CL and MTL to further improve pre-training. Although the approach enhanced performance on downstream tasks, it does not address the real-world scenario in which an MTL model should be updated to support new downstream tasks or handle distribution shifts. Moreover, the total number of sequential tasks must be known a priori for the algorithm to determine an efficient training schedule. We believe that the similarities between MTL and CL architectures could enable the construction of a CMTL model. For example, most MTL architectures in Section~\ref{subsection:arch-overview} are transformer-based MTL architectures with task-specific heads, which resemble the parameter-isolation architectures for the TIL setting in CL (Section~\ref{subsection:continual-learning}). In TIL, the task identifier is available during both training and testing and is used to identify task-specific parameters in multi-headed architectures, similar to MTL architectures \citep{ke2022continual}. This similarity can be seen in the use of adapter architectures (Section~\ref{subsubsection:arch-adapters}) in both MTL \citep{stickland2019bert, pfeiffer2020madx, he2021towards} and CL \citep{ke2021achieving, ke2021adapting}, as well as in the use of hypernetworks (Section~\ref{subsubsection:arch-hypernet}) in both MTL \citep{pilault2020conditionally, mahabadi2021parameter} and CL \citep{von2019continual, jin2021learn}.

\begin{figure}[thb]
    \centering
    \includegraphics[width=0.7\linewidth]{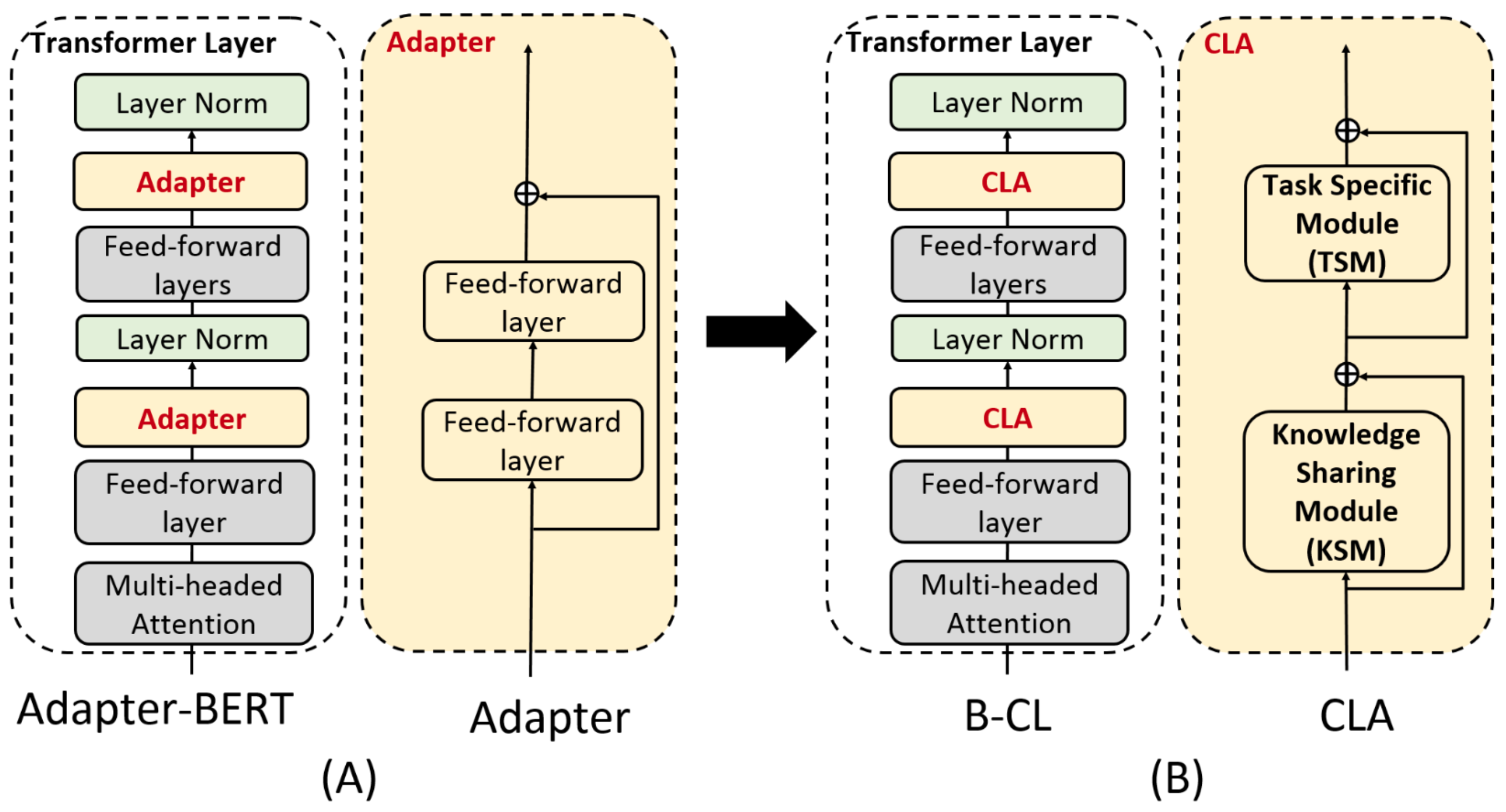}
    \caption{\textbf{(A).} Adapter-BERT \citep{houlsby2019parameter} uses adapters in a transformer layer \citep{vaswani2017attention}. Adapters are 2-layer networks with skip-connections, added twice per layer. Only adapters (yellow) and layer norm (green) are trainable while other modules (grey) are frozen. \textbf{(B).} B-CL replaces adapters with CLA, containing a knowledge-sharing module (KSM) and task-specific module (TSM), both with skip-connections. Image and modified caption are taken from \citep{ke2021adapting}.}
    \label{fig:adapter-cl}
\end{figure}

For example, if we consider the BERT layer in Figure~\ref{fig:adapter-cl}, we can observe that the adapter layers for the MTL approach in Figure~\ref{fig:adapter-cl}(A) are in the same positions as the CL adapters in Figure~\ref{fig:adapter-cl}(B) \citep{ke2021adapting}. To switch from MTL to CL we only need to change adapters. We believe that this similarity could be a promising direction for further research on the development of a CMTL model, but proper evaluation will not be possible without a good benchmark.

Finally, we believe that a benchmark that accurately represents the challenges of real-world systems could be beneficial for both researchers and practitioners to evaluate CMTL models effectively. To accomplish this, we propose combining and temporarily ordering tasks to simulate periodic re-training and CL scenarios. The benchmark can be further refined to reflect varying frequencies of periodic re-training, as well as the frequency of incoming tasks and distribution shifts between re-training periods.
Variation in these scenarios could better represent real-world situations, where a simpler CMTL model might be sufficient in some cases, while more advanced CMTL models would be needed to handle environmental changes in other situations.

\section{Conclusion}
\label{section:conclusion}
In this paper, we reviewed transformer-based MTL approaches in NLP and explored the challenges and opportunities of those approaches in the context of the ML lifecycle. We discussed how MTL can be a possible solution addressing some of the key challenges in data engineering, model development, deployment, and monitoring phases of the ML lifecycle, as compared to using multiple single-task models.

We also discussed the opportunities for applying MTL to alleviate challenges regarding model updating due to distribution shifts or evolving real-world requirements, where the ability to learn multiple tasks simultaneously can be leveraged to periodically update the model in response to changes in data and environment. However, we also acknowledged the limitations of current MTL approaches to handle updates sequentially, where CL is required. To address this, we proposed the concept of CMTL, which aims to combine the benefits of MTL and CL in a single model. We motivated creating a benchmark for the proper evaluation of CMTL models, a benchmark that better represents the challenges of production systems that can guide the development of those models.

In conclusion, MTL approaches offer many opportunities for improving the efficiency and performance of ML systems in different phases of the typical ML lifecycle. We believe that MTL will become an important part of practitioners' toolbox seeking to address the challenges in their ML systems as highlighted in this survey. 

\bibliography{paper}

\begin{thebibliography}{123}
\providecommand{\natexlab}[1]{#1}
\providecommand{\url}[1]{\texttt{#1}}
\expandafter\ifx\csname urlstyle\endcsname\relax
  \providecommand{\doi}[1]{doi: #1}\else
  \providecommand{\doi}{doi: \begingroup \urlstyle{rm}\Url}\fi

\bibitem[Abdelkader(2020)]{abdelkader2020towards}
Hala Abdelkader.
\newblock Towards robust production machine learning systems: Managing dataset
  shift.
\newblock In \emph{2020 35th IEEE/ACM International Conference on Automated
  Software Engineering (ASE)}, pp.\  1164--1166. IEEE, 2020.

\bibitem[Abhadiomhen et~al.(2022)Abhadiomhen, Nzeh, Ganaa, Nwagwu, Okereke, and
  Routray]{abhadiomhen2022supervised}
Stanley~Ebhohimhen Abhadiomhen, Royransom~Chimela Nzeh, Ernest~Domanaanmwi
  Ganaa, Honour~Chika Nwagwu, George~Emeka Okereke, and Sidheswar Routray.
\newblock Supervised shallow multi-task learning: analysis of methods.
\newblock \emph{Neural Processing Letters}, 54\penalty0 (3):\penalty0
  2491--2508, 2022.

\bibitem[Aghajanyan et~al.(2021)Aghajanyan, Gupta, Shrivastava, Chen,
  Zettlemoyer, and Gupta]{aghajanyan2021muppet}
Armen Aghajanyan, Anchit Gupta, Akshat Shrivastava, Xilun Chen, Luke
  Zettlemoyer, and Sonal Gupta.
\newblock Muppet: Massive multi-task representations with pre-finetuning.
\newblock In \emph{Proceedings of the 2021 Conference on Empirical Methods in
  Natural Language Processing}, pp.\  5799--5811, 2021.

\bibitem[Aribandi et~al.(2021)Aribandi, Tay, Schuster, Rao, Zheng, Mehta,
  Zhuang, Tran, Bahri, Ni, et~al.]{aribandiext5}
Vamsi Aribandi, Yi~Tay, Tal Schuster, Jinfeng Rao, Huaixiu~Steven Zheng,
  Sanket~Vaibhav Mehta, Honglei Zhuang, Vinh~Q Tran, Dara Bahri, Jianmo Ni,
  et~al.
\newblock Ext5: Towards extreme multi-task scaling for transfer learning.
\newblock In \emph{International Conference on Learning Representations}, 2021.

\bibitem[Ashmore et~al.(2021)Ashmore, Calinescu, and
  Paterson]{ashmore2021assuring}
Rob Ashmore, Radu Calinescu, and Colin Paterson.
\newblock Assuring the machine learning lifecycle: Desiderata, methods, and
  challenges.
\newblock \emph{ACM Computing Surveys (CSUR)}, 54\penalty0 (5):\penalty0 1--39,
  2021.

\bibitem[Bao et~al.(2020)Bao, Dong, Wei, Wang, Yang, Liu, Wang, Gao, Piao,
  Zhou, and Hon]{pmlr_v119_bao20a}
Hangbo Bao, Li~Dong, Furu Wei, Wenhui Wang, Nan Yang, Xiaodong Liu, Yu~Wang,
  Jianfeng Gao, Songhao Piao, Ming Zhou, and Hsiao-Wuen Hon.
\newblock {U}ni{LM}v2: Pseudo-masked language models for unified language model
  pre-training.
\newblock In Hal~Daumé III and Aarti Singh (eds.), \emph{Proceedings of the
  37th International Conference on Machine Learning}, volume 119 of
  \emph{Proceedings of Machine Learning Research}, pp.\  642--652. PMLR, 13--18
  Jul 2020.
\newblock URL \url{https://proceedings.mlr.press/v119/bao20a.html}.

\bibitem[Bernardi et~al.(2019)Bernardi, Mavridis, and Estevez]{bernardi2019150}
Lucas Bernardi, Themistoklis Mavridis, and Pablo Estevez.
\newblock 150 successful machine learning models: 6 lessons learned at booking.
  com.
\newblock In \emph{Proceedings of the 25th ACM SIGKDD international conference
  on knowledge discovery \& data mining}, pp.\  1743--1751, 2019.

\bibitem[Brown et~al.(2020)Brown, Mann, Ryder, Subbiah, Kaplan, Dhariwal,
  Neelakantan, Shyam, Sastry, Askell, Agarwal, Herbert{-}Voss, Krueger,
  Henighan, Child, Ramesh, Ziegler, Wu, Winter, Hesse, Chen, Sigler, Litwin,
  Gray, Chess, Clark, Berner, McCandlish, Radford, Sutskever, and
  Amodei]{brown2020language}
Tom~B. Brown, Benjamin Mann, Nick Ryder, Melanie Subbiah, Jared Kaplan,
  Prafulla Dhariwal, Arvind Neelakantan, Pranav Shyam, Girish Sastry, Amanda
  Askell, Sandhini Agarwal, Ariel Herbert{-}Voss, Gretchen Krueger, Tom
  Henighan, Rewon Child, Aditya Ramesh, Daniel~M. Ziegler, Jeffrey Wu, Clemens
  Winter, Christopher Hesse, Mark Chen, Eric Sigler, Mateusz Litwin, Scott
  Gray, Benjamin Chess, Jack Clark, Christopher Berner, Sam McCandlish, Alec
  Radford, Ilya Sutskever, and Dario Amodei.
\newblock Language models are few-shot learners.
\newblock \emph{CoRR}, abs/2005.14165, 2020.
\newblock URL \url{https://arxiv.org/abs/2005.14165}.

\bibitem[Caruana(1997)]{caruana1997multitask}
Rich Caruana.
\newblock Multitask learning.
\newblock \emph{Machine learning}, 28\penalty0 (1):\penalty0 41--75, 1997.

\bibitem[Chen et~al.(2021)Chen, Zhang, and Yang]{chen2021multi}
Shijie Chen, Yu~Zhang, and Qiang Yang.
\newblock Multi-task learning in natural language processing: An overview.
\newblock \emph{arXiv preprint arXiv:2109.09138}, 2021.

\bibitem[Chen \& Liu(2018)Chen and Liu]{chen2018lifelong}
Zhiyuan Chen and Bing Liu.
\newblock Lifelong machine learning.
\newblock \emph{Synthesis Lectures on Artificial Intelligence and Machine
  Learning}, 12\penalty0 (3):\penalty0 1--207, 2018.

\bibitem[Chui et~al.(2022)Chui, Hall, Mayhew, and Singla]{chui2022stateofai}
Michael Chui, Bryce Hall, Helen Mayhew, and Alex Singla.
\newblock The state of ai in 2022--and a half decade in review, Dec 2022.
\newblock URL
  \url{https://www.mckinsey.com/capabilities/quantumblack/our-insights/the-state-of-ai-in-2022-and-a-half-decade-in-review}.
\newblock Accessed: 2023-01-15.

\bibitem[Chung et~al.(2022)Chung, Hou, Longpre, Zoph, Tay, Fedus, Li, Wang,
  Dehghani, Brahma, et~al.]{chung2022scaling}
Hyung~Won Chung, Le~Hou, Shayne Longpre, Barret Zoph, Yi~Tay, William Fedus,
  Eric Li, Xuezhi Wang, Mostafa Dehghani, Siddhartha Brahma, et~al.
\newblock Scaling instruction-finetuned language models.
\newblock \emph{arXiv preprint arXiv:2210.11416}, 2022.

\bibitem[Clark et~al.(2019{\natexlab{a}})Clark, Yatskar, and
  Zettlemoyer]{clark2019dont}
Christopher Clark, Mark Yatskar, and Luke Zettlemoyer.
\newblock Don{'}t take the easy way out: Ensemble based methods for avoiding
  known dataset biases.
\newblock In \emph{Proceedings of the 2019 Conference on Empirical Methods in
  Natural Language Processing and the 9th International Joint Conference on
  Natural Language Processing (EMNLP-IJCNLP)}, pp.\  4069--4082, Hong Kong,
  China, November 2019{\natexlab{a}}. Association for Computational
  Linguistics.
\newblock \doi{10.18653/v1/D19-1418}.
\newblock URL \url{https://aclanthology.org/D19-1418}.

\bibitem[Clark et~al.(2019{\natexlab{b}})Clark, Luong, Khandelwal, Manning, and
  Le]{clark2019bam}
Kevin Clark, Minh{-}Thang Luong, Urvashi Khandelwal, Christopher~D. Manning,
  and Quoc~V. Le.
\newblock Bam! born-again multi-task networks for natural language
  understanding.
\newblock \emph{CoRR}, abs/1907.04829, 2019{\natexlab{b}}.
\newblock URL \url{http://arxiv.org/abs/1907.04829}.

\bibitem[Crawshaw(2020)]{crawshaw2020multi}
Michael Crawshaw.
\newblock Multi-task learning with deep neural networks: A survey.
\newblock \emph{arXiv preprint arXiv:2009.09796}, 2020.

\bibitem[De~Lange et~al.(2021)De~Lange, Aljundi, Masana, Parisot, Jia,
  Leonardis, Slabaugh, and Tuytelaars]{de2021continual}
Matthias De~Lange, Rahaf Aljundi, Marc Masana, Sarah Parisot, Xu~Jia,
  Ale{\v{s}} Leonardis, Gregory Slabaugh, and Tinne Tuytelaars.
\newblock A continual learning survey: Defying forgetting in classification
  tasks.
\newblock \emph{IEEE transactions on pattern analysis and machine
  intelligence}, 44\penalty0 (7):\penalty0 3366--3385, 2021.

\bibitem[Devlin et~al.(2019)Devlin, Chang, Lee, and
  Toutanova]{devlin-etal-2019-bert}
Jacob Devlin, Ming-Wei Chang, Kenton Lee, and Kristina Toutanova.
\newblock {BERT}: Pre-training of deep bidirectional transformers for language
  understanding.
\newblock In \emph{Proceedings of the 2019 Conference of the North {A}merican
  Chapter of the Association for Computational Linguistics: Human Language
  Technologies, Volume 1 (Long and Short Papers)}, pp.\  4171--4186,
  Minneapolis, Minnesota, June 2019. Association for Computational Linguistics.
\newblock \doi{10.18653/v1/N19-1423}.
\newblock URL \url{https://aclanthology.org/N19-1423}.

\bibitem[Ding et~al.(2018)Ding, Tarokh, and Yang]{ding2018model}
Jie Ding, Vahid Tarokh, and Yuhong Yang.
\newblock Model selection techniques: An overview.
\newblock \emph{IEEE Signal Processing Magazine}, 35\penalty0 (6):\penalty0
  16--34, 2018.

\bibitem[Ditzler et~al.(2015)Ditzler, Roveri, Alippi, and
  Polikar]{ditzler2015learning}
Gregory Ditzler, Manuel Roveri, Cesare Alippi, and Robi Polikar.
\newblock Learning in nonstationary environments: A survey.
\newblock \emph{IEEE Computational Intelligence Magazine}, 10\penalty0
  (4):\penalty0 12--25, 2015.

\bibitem[González-Garduño \& Søgaard(2018)González-Garduño and
  Søgaard]{González_Garduño_Søgaard_2018}
Ana González-Garduño and Anders Søgaard.
\newblock Learning to predict readability using eye-movement data from natives
  and learners.
\newblock \emph{Proceedings of the AAAI Conference on Artificial Intelligence},
  32\penalty0 (1), Apr. 2018.
\newblock \doi{10.1609/aaai.v32i1.11978}.
\newblock URL \url{https://ojs.aaai.org/index.php/AAAI/article/view/11978}.

\bibitem[Guo et~al.(2019)Guo, Pasunuru, and Bansal]{guo2019autosem}
Han Guo, Ramakanth Pasunuru, and Mohit Bansal.
\newblock Autosem: Automatic task selection and mixing in multi-task learning.
\newblock \emph{CoRR}, abs/1904.04153, 2019.
\newblock URL \url{http://arxiv.org/abs/1904.04153}.

\bibitem[Guo et~al.(2018)Guo, Haque, Huang, Yeung, and Fei-Fei]{Guo_2018_ECCV}
Michelle Guo, Albert Haque, De-An Huang, Serena Yeung, and Li~Fei-Fei.
\newblock Dynamic task prioritization for multitask learning.
\newblock In \emph{Proceedings of the European Conference on Computer Vision
  (ECCV)}, September 2018.

\bibitem[Gupta \& Agrawal(2022)Gupta and Agrawal]{gupta2022compression}
Manish Gupta and Puneet Agrawal.
\newblock Compression of deep learning models for text: A survey.
\newblock \emph{ACM Transactions on Knowledge Discovery from Data (TKDD)},
  16\penalty0 (4):\penalty0 1--55, 2022.

\bibitem[Ha et~al.(2016)Ha, Dai, and Le]{ha2016hypernetworks}
David Ha, Andrew Dai, and Quoc~V. Le.
\newblock Hypernetworks, 2016.
\newblock URL \url{https://arxiv.org/abs/1609.09106}.

\bibitem[Haldar et~al.(2019)Haldar, Abdool, Ramanathan, Xu, Yang, Duan, Zhang,
  Barrow-Williams, Turnbull, Collins, et~al.]{haldar2019applying}
Malay Haldar, Mustafa Abdool, Prashant Ramanathan, Tao Xu, Shulin Yang,
  Huizhong Duan, Qing Zhang, Nick Barrow-Williams, Bradley~C Turnbull,
  Brendan~M Collins, et~al.
\newblock Applying deep learning to airbnb search.
\newblock In \emph{proceedings of the 25th ACM SIGKDD international conference
  on knowledge discovery \& Data Mining}, pp.\  1927--1935, 2019.

\bibitem[Han et~al.(2021)Han, Zhao, Ding, Liu, and Sun]{han2021ptr}
Xu~Han, Weilin Zhao, Ning Ding, Zhiyuan Liu, and Maosong Sun.
\newblock {PTR:} prompt tuning with rules for text classification.
\newblock \emph{CoRR}, abs/2105.11259, 2021.
\newblock URL \url{https://arxiv.org/abs/2105.11259}.

\bibitem[He et~al.(2021)He, Zhou, Ma, Berg-Kirkpatrick, and
  Neubig]{he2021towards}
Junxian He, Chunting Zhou, Xuezhe Ma, Taylor Berg-Kirkpatrick, and Graham
  Neubig.
\newblock Towards a unified view of parameter-efficient transfer learning.
\newblock \emph{arXiv preprint arXiv:2110.04366}, 2021.

\bibitem[He et~al.(2022)He, Zheng, Tay, Gupta, Du, Aribandi, Zhao, Li, Chen,
  Metzler, Cheng, and Chi]{he2022hyperprompt}
Yun He, Steven Zheng, Yi~Tay, Jai Gupta, Yu~Du, Vamsi Aribandi, Zhe Zhao,
  Yaguang Li, Zhao Chen, Donald Metzler, Heng-Tze Cheng, and Ed~H. Chi.
\newblock {H}yper{P}rompt: Prompt-based task-conditioning of transformers.
\newblock In Kamalika Chaudhuri, Stefanie Jegelka, Le~Song, Csaba Szepesvari,
  Gang Niu, and Sivan Sabato (eds.), \emph{Proceedings of the 39th
  International Conference on Machine Learning}, volume 162 of
  \emph{Proceedings of Machine Learning Research}, pp.\  8678--8690. PMLR,
  17--23 Jul 2022.
\newblock URL \url{https://proceedings.mlr.press/v162/he22f.html}.

\bibitem[Houlsby et~al.(2019)Houlsby, Giurgiu, Jastrzebski, Morrone,
  De~Laroussilhe, Gesmundo, Attariyan, and Gelly]{houlsby2019parameter}
Neil Houlsby, Andrei Giurgiu, Stanislaw Jastrzebski, Bruna Morrone, Quentin
  De~Laroussilhe, Andrea Gesmundo, Mona Attariyan, and Sylvain Gelly.
\newblock Parameter-efficient transfer learning for nlp.
\newblock In \emph{International Conference on Machine Learning}, pp.\
  2790--2799. PMLR, 2019.

\bibitem[Hsu et~al.(2018)Hsu, Liu, Ramasamy, and Kira]{hsu2018re}
Yen-Chang Hsu, Yen-Cheng Liu, Anita Ramasamy, and Zsolt Kira.
\newblock Re-evaluating continual learning scenarios: A categorization and case
  for strong baselines.
\newblock \emph{arXiv preprint arXiv:1810.12488}, 2018.

\bibitem[Hu et~al.(2021)Hu, Shen, Wallis, Allen-Zhu, Li, Wang, Wang, and
  Chen]{hu2021lora}
Edward~J. Hu, Yelong Shen, Phillip Wallis, Zeyuan Allen-Zhu, Yuanzhi Li, Shean
  Wang, Lu~Wang, and Weizhu Chen.
\newblock Lora: Low-rank adaptation of large language models, 2021.
\newblock URL \url{https://arxiv.org/abs/2106.09685}.

\bibitem[Huyen(2022)]{huyen2022designing}
Chip Huyen.
\newblock \emph{Designing Machine Learning Systems}.
\newblock " O'Reilly Media, Inc.", 2022.

\bibitem[Jean et~al.(2019)Jean, Firat, and Johnson]{Jean2019adaptive}
S{\'{e}}bastien Jean, Orhan Firat, and Melvin Johnson.
\newblock Adaptive scheduling for multi-task learning.
\newblock \emph{CoRR}, abs/1909.06434, 2019.
\newblock URL \url{http://arxiv.org/abs/1909.06434}.

\bibitem[Jin et~al.(2021)Jin, Lin, Rostami, and Ren]{jin2021learn}
Xisen Jin, Bill~Yuchen Lin, Mohammad Rostami, and Xiang Ren.
\newblock Learn continually, generalize rapidly: lifelong knowledge
  accumulation for few-shot learning.
\newblock \emph{arXiv preprint arXiv:2104.08808}, 2021.

\bibitem[Kaiser et~al.(2017)Kaiser, Gomez, Shazeer, Vaswani, Parmar, Jones, and
  Uszkoreit]{kaiser2017one}
Lukasz Kaiser, Aidan~N. Gomez, Noam Shazeer, Ashish Vaswani, Niki Parmar, Llion
  Jones, and Jakob Uszkoreit.
\newblock One model to learn them all, 2017.
\newblock URL \url{https://arxiv.org/abs/1706.05137}.

\bibitem[Karpathy(2021)]{tesla-2021-ai-day}
Andrej Karpathy.
\newblock {HydraNets - Tesla AI Day 2021}, 8 2021.
\newblock URL
  \url{https://www.youtube.com/watch?t=4284&v=j0z4FweCy4M&feature=youtu.be}.
\newblock Accessed: 2023-02-15.

\bibitem[Ke \& Liu(2022)Ke and Liu]{ke2022continual}
Zixuan Ke and Bing Liu.
\newblock Continual learning of natural language processing tasks: A survey.
\newblock \emph{arXiv preprint arXiv:2211.12701}, 2022.

\bibitem[Ke et~al.(2021{\natexlab{a}})Ke, Liu, Ma, Xu, and
  Shu]{ke2021achieving}
Zixuan Ke, Bing Liu, Nianzu Ma, Hu~Xu, and Lei Shu.
\newblock Achieving forgetting prevention and knowledge transfer in continual
  learning.
\newblock \emph{Advances in Neural Information Processing Systems},
  34:\penalty0 22443--22456, 2021{\natexlab{a}}.

\bibitem[Ke et~al.(2021{\natexlab{b}})Ke, Xu, and Liu]{ke2021adapting}
Zixuan Ke, Hu~Xu, and Bing Liu.
\newblock Adapting bert for continual learning of a sequence of aspect
  sentiment classification tasks.
\newblock \emph{arXiv preprint arXiv:2112.03271}, 2021{\natexlab{b}}.

\bibitem[Kendall et~al.(2018)Kendall, Gal, and Cipolla]{Kendall2018CVPR}
Alex Kendall, Yarin Gal, and Roberto Cipolla.
\newblock Multi-task learning using uncertainty to weigh losses for scene
  geometry and semantics.
\newblock In \emph{Proceedings of the IEEE Conference on Computer Vision and
  Pattern Recognition (CVPR)}, June 2018.

\bibitem[Kim \& Doshi-Velez(2021)Kim and Doshi-Velez]{kim2021machine}
Been Kim and Finale Doshi-Velez.
\newblock Machine learning techniques for accountability.
\newblock \emph{AI Magazine}, 42\penalty0 (1):\penalty0 47--52, 2021.

\bibitem[Kirstein et~al.(2022)Kirstein, Wahle, Ruas, and
  Gipp]{kirstein2022analyzing}
Frederic Kirstein, Jan~Philip Wahle, Terry Ruas, and Bela Gipp.
\newblock Analyzing multi-task learning for abstractive text summarization.
\newblock \emph{arXiv preprint arXiv:2210.14606}, 2022.

\bibitem[Lakshmanan et~al.(2020)Lakshmanan, Robinson, and
  Munn]{lakshmanan2020machine}
Valliappa Lakshmanan, Sara Robinson, and Michael Munn.
\newblock \emph{Machine learning design patterns}.
\newblock O'Reilly Media, 2020.

\bibitem[Lee et~al.(2018)Lee, Yang, and Hwang]{lee2018deep}
Hae~Beom Lee, Eunho Yang, and Sung~Ju Hwang.
\newblock Deep asymmetric multi-task feature learning.
\newblock In Jennifer Dy and Andreas Krause (eds.), \emph{Proceedings of the
  35th International Conference on Machine Learning}, volume~80 of
  \emph{Proceedings of Machine Learning Research}, pp.\  2956--2964. PMLR,
  10--15 Jul 2018.
\newblock URL \url{https://proceedings.mlr.press/v80/lee18d.html}.

\bibitem[Lester et~al.(2021)Lester, Al{-}Rfou, and Constant]{lester2021power}
Brian Lester, Rami Al{-}Rfou, and Noah Constant.
\newblock The power of scale for parameter-efficient prompt tuning.
\newblock \emph{CoRR}, abs/2104.08691, 2021.
\newblock URL \url{https://arxiv.org/abs/2104.08691}.

\bibitem[Lewis et~al.(2020)Lewis, Liu, Goyal, Ghazvininejad, Mohamed, Levy,
  Stoyanov, and Zettlemoyer]{lewis2020bart}
Mike Lewis, Yinhan Liu, Naman Goyal, Marjan Ghazvininejad, Abdelrahman Mohamed,
  Omer Levy, Veselin Stoyanov, and Luke Zettlemoyer.
\newblock {BART}: Denoising sequence-to-sequence pre-training for natural
  language generation, translation, and comprehension.
\newblock In \emph{Proceedings of the 58th Annual Meeting of the Association
  for Computational Linguistics}, pp.\  7871--7880, Online, July 2020.
  Association for Computational Linguistics.
\newblock \doi{10.18653/v1/2020.acl-main.703}.
\newblock URL \url{https://aclanthology.org/2020.acl-main.703}.

\bibitem[Li \& Liang(2021)Li and Liang]{li2021prefix}
Xiang~Lisa Li and Percy Liang.
\newblock Prefix-tuning: Optimizing continuous prompts for generation.
\newblock \emph{arXiv preprint arXiv:2101.00190}, 2021.

\bibitem[Lin et~al.(2019)Lin, Zhen, Li, Zhang, and Kwong]{lin2019pareto}
Xi~Lin, Hui-Ling Zhen, Zhenhua Li, Qing-Fu Zhang, and Sam Kwong.
\newblock Pareto multi-task learning.
\newblock In H.~Wallach, H.~Larochelle, A.~Beygelzimer, F.~d\textquotesingle
  Alch\'{e}-Buc, E.~Fox, and R.~Garnett (eds.), \emph{Advances in Neural
  Information Processing Systems}, volume~32. Curran Associates, Inc., 2019.
\newblock URL
  \url{https://proceedings.neurips.cc/paper/2019/file/685bfde03eb646c27ed565881917c71c-Paper.pdf}.

\bibitem[Liu et~al.(2021{\natexlab{a}})Liu, Yuan, Fu, Jiang, Hayashi, and
  Neubig]{liu2021pretrain}
Pengfei Liu, Weizhe Yuan, Jinlan Fu, Zhengbao Jiang, Hiroaki Hayashi, and
  Graham Neubig.
\newblock Pre-train, prompt, and predict: {A} systematic survey of prompting
  methods in natural language processing.
\newblock \emph{CoRR}, abs/2107.13586, 2021{\natexlab{a}}.
\newblock URL \url{https://arxiv.org/abs/2107.13586}.

\bibitem[Liu et~al.(2019{\natexlab{a}})Liu, Liang, and
  Gitter]{Liu_Liang_Gitter_2019}
Shengchao Liu, Yingyu Liang, and Anthony Gitter.
\newblock Loss-balanced task weighting to reduce negative transfer in
  multi-task learning.
\newblock \emph{Proceedings of the AAAI Conference on Artificial Intelligence},
  33\penalty0 (01):\penalty0 9977--9978, Jul. 2019{\natexlab{a}}.
\newblock \doi{10.1609/aaai.v33i01.33019977}.
\newblock URL \url{https://ojs.aaai.org/index.php/AAAI/article/view/5125}.

\bibitem[Liu et~al.(2021{\natexlab{b}})Liu, Zheng, Du, Ding, Qian, Yang, and
  Tang]{liu2021gpt}
Xiao Liu, Yanan Zheng, Zhengxiao Du, Ming Ding, Yujie Qian, Zhilin Yang, and
  Jie Tang.
\newblock {GPT} understands, too.
\newblock \emph{CoRR}, abs/2103.10385, 2021{\natexlab{b}}.
\newblock URL \url{https://arxiv.org/abs/2103.10385}.

\bibitem[Liu et~al.(2019{\natexlab{b}})Liu, He, Chen, and
  Gao]{liu2019multitask}
Xiaodong Liu, Pengcheng He, Weizhu Chen, and Jianfeng Gao.
\newblock Multi-task deep neural networks for natural language understanding,
  2019{\natexlab{b}}.
\newblock URL \url{https://arxiv.org/abs/1901.11504}.

\bibitem[Liu et~al.(2019{\natexlab{c}})Liu, Ott, Goyal, Du, Joshi, Chen, Levy,
  Lewis, Zettlemoyer, and Stoyanov]{liu2019roberta}
Yinhan Liu, Myle Ott, Naman Goyal, Jingfei Du, Mandar Joshi, Danqi Chen, Omer
  Levy, Mike Lewis, Luke Zettlemoyer, and Veselin Stoyanov.
\newblock Roberta: {A} robustly optimized {BERT} pretraining approach.
\newblock \emph{CoRR}, abs/1907.11692, 2019{\natexlab{c}}.
\newblock URL \url{http://arxiv.org/abs/1907.11692}.

\bibitem[Long et~al.(2017)Long, CAO, Wang, and Yu]{long2017learning}
Mingsheng Long, ZHANGJIE CAO, Jianmin Wang, and Philip~S Yu.
\newblock Learning multiple tasks with multilinear relationship networks.
\newblock In I.~Guyon, U.~Von Luxburg, S.~Bengio, H.~Wallach, R.~Fergus,
  S.~Vishwanathan, and R.~Garnett (eds.), \emph{Advances in Neural Information
  Processing Systems}, volume~30. Curran Associates, Inc., 2017.
\newblock URL
  \url{https://proceedings.neurips.cc/paper/2017/file/03e0704b5690a2dee1861dc3ad3316c9-Paper.pdf}.

\bibitem[Lopez-Paz \& Ranzato(2017)Lopez-Paz and
  Ranzato]{lopez_paz2017gradient}
David Lopez-Paz and Marc\textquotesingle~Aurelio Ranzato.
\newblock Gradient episodic memory for continual learning.
\newblock In I.~Guyon, U.~Von Luxburg, S.~Bengio, H.~Wallach, R.~Fergus,
  S.~Vishwanathan, and R.~Garnett (eds.), \emph{Advances in Neural Information
  Processing Systems}, volume~30. Curran Associates, Inc., 2017.
\newblock URL
  \url{https://proceedings.neurips.cc/paper/2017/file/f87522788a2be2d171666752f97ddebb-Paper.pdf}.

\bibitem[Mahabadi et~al.(2021)Mahabadi, Ruder, Dehghani, and
  Henderson]{mahabadi2021parameter}
Rabeeh~Karimi Mahabadi, Sebastian Ruder, Mostafa Dehghani, and James Henderson.
\newblock Parameter-efficient multi-task fine-tuning for transformers via
  shared hypernetworks, 2021.
\newblock URL \url{https://arxiv.org/abs/2106.04489}.

\bibitem[Marchant(2011)]{marchant2011growing}
Gary~E Marchant.
\newblock \emph{The growing gap between emerging technologies and the law}.
\newblock Springer, 2011.

\bibitem[McCann et~al.(2018)McCann, Keskar, Xiong, and
  Socher]{mccann2018natural}
Bryan McCann, Nitish~Shirish Keskar, Caiming Xiong, and Richard Socher.
\newblock The natural language decathlon: Multitask learning as question
  answering.
\newblock \emph{arXiv preprint arXiv:1806.08730}, 2018.

\bibitem[Mehrabi et~al.(2021)Mehrabi, Morstatter, Saxena, Lerman, and
  Galstyan]{mehrabi2021survey}
Ninareh Mehrabi, Fred Morstatter, Nripsuta Saxena, Kristina Lerman, and Aram
  Galstyan.
\newblock A survey on bias and fairness in machine learning.
\newblock \emph{ACM Computing Surveys (CSUR)}, 54\penalty0 (6):\penalty0 1--35,
  2021.

\bibitem[Nahar et~al.(2022)Nahar, Zhou, Lewis, and
  K{\"a}stner]{nahar2022collaboration}
Nadia Nahar, Shurui Zhou, Grace Lewis, and Christian K{\"a}stner.
\newblock Collaboration challenges in building ml-enabled systems:
  Communication, documentation, engineering, and process.
\newblock In \emph{Proceedings of the 44th International Conference on Software
  Engineering}, pp.\  413--425, 2022.

\bibitem[Pacheco et~al.(2018)Pacheco, Exposito, Gineste, Baudoin, and
  Aguilar]{pacheco2018towards}
Fannia Pacheco, Ernesto Exposito, Mathieu Gineste, Cedric Baudoin, and Jose
  Aguilar.
\newblock Towards the deployment of machine learning solutions in network
  traffic classification: A systematic survey.
\newblock \emph{IEEE Communications Surveys \& Tutorials}, 21\penalty0
  (2):\penalty0 1988--2014, 2018.

\bibitem[Paleyes et~al.(2022)Paleyes, Urma, and Lawrence]{paleyes2022survey}
Andrei Paleyes, Raoul-Gabriel Urma, and Neil~D Lawrence.
\newblock Challenges in deploying machine learning: a survey of case studies.
\newblock \emph{ACM Computing Surveys}, 55\penalty0 (6):\penalty0 1--29, 2022.

\bibitem[Parmar et~al.(2022)Parmar, Mishra, Purohit, Luo, Mohammad, and
  Baral]{parmar2022inboxbart}
Mihir Parmar, Swaroop Mishra, Mirali Purohit, Man Luo, Murad Mohammad, and
  Chitta Baral.
\newblock In-{B}o{XBART}: Get instructions into biomedical multi-task learning.
\newblock In \emph{Findings of the Association for Computational Linguistics:
  NAACL 2022}, pp.\  112--128, Seattle, United States, July 2022. Association
  for Computational Linguistics.
\newblock \doi{10.18653/v1/2022.findings-naacl.10}.
\newblock URL \url{https://aclanthology.org/2022.findings-naacl.10}.

\bibitem[Pascal et~al.(2021)Pascal, Michiardi, Bost, Huet, and
  Zuluaga]{Pascal_Michiardi_Bost_Huet_Zuluaga_2021}
Lucas Pascal, Pietro Michiardi, Xavier Bost, Benoit Huet, and Maria~A. Zuluaga.
\newblock Maximum roaming multi-task learning.
\newblock \emph{Proceedings of the AAAI Conference on Artificial Intelligence},
  35\penalty0 (10):\penalty0 9331--9341, May 2021.
\newblock \doi{10.1609/aaai.v35i10.17125}.
\newblock URL \url{https://ojs.aaai.org/index.php/AAAI/article/view/17125}.

\bibitem[Pei et~al.(2022)Pei, Liu, Wang, and Wang]{pei2022requirements}
Zhongyi Pei, Lin Liu, Chen Wang, and Jianmin Wang.
\newblock Requirements engineering for machine learning: A review and
  reflection.
\newblock In \emph{2022 IEEE 30th International Requirements Engineering
  Conference Workshops (REW)}, pp.\  166--175. IEEE, 2022.

\bibitem[Petroni et~al.(2019)Petroni, Rockt{\"a}schel, Riedel, Lewis, Bakhtin,
  Wu, and Miller]{petroni2019language}
Fabio Petroni, Tim Rockt{\"a}schel, Sebastian Riedel, Patrick Lewis, Anton
  Bakhtin, Yuxiang Wu, and Alexander Miller.
\newblock Language models as knowledge bases?
\newblock In \emph{Proceedings of the 2019 Conference on Empirical Methods in
  Natural Language Processing and the 9th International Joint Conference on
  Natural Language Processing (EMNLP-IJCNLP)}, pp.\  2463--2473, Hong Kong,
  China, November 2019. Association for Computational Linguistics.
\newblock \doi{10.18653/v1/D19-1250}.
\newblock URL \url{https://aclanthology.org/D19-1250}.

\bibitem[Pfeiffer et~al.(2020{\natexlab{a}})Pfeiffer, Kamath, R{\"u}ckl{\'e},
  Cho, and Gurevych]{pfeiffer2020adapterfusion}
Jonas Pfeiffer, Aishwarya Kamath, Andreas R{\"u}ckl{\'e}, Kyunghyun Cho, and
  Iryna Gurevych.
\newblock Adapterfusion: Non-destructive task composition for transfer
  learning.
\newblock \emph{arXiv preprint arXiv:2005.00247}, 2020{\natexlab{a}}.

\bibitem[Pfeiffer et~al.(2020{\natexlab{b}})Pfeiffer, R{\"u}ckl{\'e}, Poth,
  Kamath, Vuli{\'c}, Ruder, Cho, and Gurevych]{pfeiffer2020adapterhub}
Jonas Pfeiffer, Andreas R{\"u}ckl{\'e}, Clifton Poth, Aishwarya Kamath, Ivan
  Vuli{\'c}, Sebastian Ruder, Kyunghyun Cho, and Iryna Gurevych.
\newblock Adapterhub: A framework for adapting transformers.
\newblock \emph{arXiv preprint arXiv:2007.07779}, 2020{\natexlab{b}}.

\bibitem[Pfeiffer et~al.(2020{\natexlab{c}})Pfeiffer, Vulic, Gurevych, and
  Ruder]{pfeiffer2020madx}
Jonas Pfeiffer, Ivan Vulic, Iryna Gurevych, and Sebastian Ruder.
\newblock {MAD-X:} an adapter-based framework for multi-task cross-lingual
  transfer.
\newblock \emph{CoRR}, abs/2005.00052, 2020{\natexlab{c}}.
\newblock URL \url{https://arxiv.org/abs/2005.00052}.

\bibitem[Pilault et~al.(2020)Pilault, Elhattami, and
  Pal]{pilault2020conditionally}
Jonathan Pilault, Amine Elhattami, and Christopher Pal.
\newblock Conditionally adaptive multi-task learning: Improving transfer
  learning in nlp using fewer parameters \& less data, 2020.
\newblock URL \url{https://arxiv.org/abs/2009.09139}.

\bibitem[Politou et~al.(2018)Politou, Alepis, and
  Patsakis]{politou2018forgetting}
Eugenia Politou, Efthimios Alepis, and Constantinos Patsakis.
\newblock Forgetting personal data and revoking consent under the gdpr:
  Challenges and proposed solutions.
\newblock \emph{Journal of cybersecurity}, 4\penalty0 (1):\penalty0 tyy001,
  2018.

\bibitem[Polyzotis et~al.(2018)Polyzotis, Roy, Whang, and
  Zinkevich]{polyzotis2018data}
Neoklis Polyzotis, Sudip Roy, Steven~Euijong Whang, and Martin Zinkevich.
\newblock Data lifecycle challenges in production machine learning: a survey.
\newblock \emph{ACM SIGMOD Record}, 47\penalty0 (2):\penalty0 17--28, 2018.

\bibitem[Pramanik et~al.(2019)Pramanik, Agrawal, and
  Hussain]{subhojeet2019omninet}
Subhojeet Pramanik, Priyanka Agrawal, and Aman Hussain.
\newblock Omninet: {A} unified architecture for multi-modal multi-task
  learning.
\newblock \emph{CoRR}, abs/1907.07804, 2019.
\newblock URL \url{http://arxiv.org/abs/1907.07804}.

\bibitem[Radford et~al.(2019)Radford, Wu, Child, Luan, Amodei, Sutskever,
  et~al.]{radford2019gpt2}
Alec Radford, Jeffrey Wu, Rewon Child, David Luan, Dario Amodei, Ilya
  Sutskever, et~al.
\newblock Language models are unsupervised multitask learners.
\newblock \emph{OpenAI blog}, 1\penalty0 (8):\penalty0 9, 2019.

\bibitem[Raffel et~al.(2020)Raffel, Shazeer, Roberts, Lee, Narang, Matena,
  Zhou, Li, Liu, et~al.]{raffel2020exploring}
Colin Raffel, Noam Shazeer, Adam Roberts, Katherine Lee, Sharan Narang, Michael
  Matena, Yanqi Zhou, Wei Li, Peter~J Liu, et~al.
\newblock Exploring the limits of transfer learning with a unified text-to-text
  transformer.
\newblock \emph{J. Mach. Learn. Res.}, 21\penalty0 (140):\penalty0 1--67, 2020.

\bibitem[Rebuffi et~al.(2017)Rebuffi, Bilen, and Vedaldi]{rebuffi2017learning}
Sylvestre-Alvise Rebuffi, Hakan Bilen, and Andrea Vedaldi.
\newblock Learning multiple visual domains with residual adapters.
\newblock \emph{Advances in neural information processing systems}, 30, 2017.

\bibitem[Ren et~al.(2020)Ren, Zheng, Qin, and Liu]{ren2020adversarial}
Kui Ren, Tianhang Zheng, Zhan Qin, and Xue Liu.
\newblock Adversarial attacks and defenses in deep learning.
\newblock \emph{Engineering}, 6\penalty0 (3):\penalty0 346--360, 2020.

\bibitem[Renggli et~al.(2019)Renggli, Karla{\v{s}}, Ding, Liu, Schawinski, Wu,
  and Zhang]{renggli2019continuous}
Cedric Renggli, Bojan Karla{\v{s}}, Bolin Ding, Feng Liu, Kevin Schawinski,
  Wentao Wu, and Ce~Zhang.
\newblock Continuous integration of machine learning models with ease. ml/ci:
  Towards a rigorous yet practical treatment.
\newblock \emph{Proceedings of Machine Learning and Systems}, 1:\penalty0
  322--333, 2019.

\bibitem[Rosenberg et~al.(2021)Rosenberg, Shabtai, Elovici, and
  Rokach]{rosenberg2021adversarial}
Ishai Rosenberg, Asaf Shabtai, Yuval Elovici, and Lior Rokach.
\newblock Adversarial machine learning attacks and defense methods in the cyber
  security domain.
\newblock \emph{ACM Computing Surveys (CSUR)}, 54\penalty0 (5):\penalty0 1--36,
  2021.

\bibitem[Ruder(2017)]{ruder2017overview}
Sebastian Ruder.
\newblock An overview of multi-task learning in deep neural networks.
\newblock \emph{arXiv preprint arXiv:1706.05098}, 2017.

\bibitem[Ruder et~al.(2017)Ruder, Bingel, Augenstein, and
  S{\o}gaard]{ruder2017sluice}
Sebastian Ruder, Joachim Bingel, Isabelle Augenstein, and Anders S{\o}gaard.
\newblock Sluice networks: Learning what to share between loosely related
  tasks.
\newblock \emph{arXiv preprint arXiv:1705.08142}, 2, 2017.

\bibitem[Samant et~al.(2022)Samant, Bachute, Gite, and
  Kotecha]{samant2022framework}
Rahul~Manohar Samant, Mrinal Bachute, Shilpa Gite, and Ketan Kotecha.
\newblock Framework for deep learning-based language models using multi-task
  learning in natural language understanding: A systematic literature review
  and future directions.
\newblock \emph{IEEE Access}, 2022.

\bibitem[Sambasivan et~al.(2021)Sambasivan, Kapania, Highfill, Akrong,
  Paritosh, and Aroyo]{sambasivan2021everyone}
Nithya Sambasivan, Shivani Kapania, Hannah Highfill, Diana Akrong, Praveen
  Paritosh, and Lora~M Aroyo.
\newblock “everyone wants to do the model work, not the data work”: Data
  cascades in high-stakes ai.
\newblock In \emph{proceedings of the 2021 CHI Conference on Human Factors in
  Computing Systems}, pp.\  1--15, 2021.

\bibitem[Sanh et~al.(2021)Sanh, Webson, Raffel, Bach, Sutawika, Alyafeai,
  Chaffin, Stiegler, Scao, Raja, Dey, Bari, Xu, Thakker, Sharma, Szczechla,
  Kim, Chhablani, Nayak, Datta, Chang, Jiang, Wang, Manica, Shen, Yong, Pandey,
  Bawden, Wang, Neeraj, Rozen, Sharma, Santilli, Fevry, Fries, Teehan, Bers,
  Biderman, Gao, Wolf, and Rush]{sanh2021multitask}
Victor Sanh, Albert Webson, Colin Raffel, Stephen~H. Bach, Lintang Sutawika,
  Zaid Alyafeai, Antoine Chaffin, Arnaud Stiegler, Teven~Le Scao, Arun Raja,
  Manan Dey, M~Saiful Bari, Canwen Xu, Urmish Thakker, Shanya~Sharma Sharma,
  Eliza Szczechla, Taewoon Kim, Gunjan Chhablani, Nihal Nayak, Debajyoti Datta,
  Jonathan Chang, Mike Tian-Jian Jiang, Han Wang, Matteo Manica, Sheng Shen,
  Zheng~Xin Yong, Harshit Pandey, Rachel Bawden, Thomas Wang, Trishala Neeraj,
  Jos Rozen, Abheesht Sharma, Andrea Santilli, Thibault Fevry, Jason~Alan
  Fries, Ryan Teehan, Tali Bers, Stella Biderman, Leo Gao, Thomas Wolf, and
  Alexander~M. Rush.
\newblock Multitask prompted training enables zero-shot task generalization,
  2021.
\newblock URL \url{https://arxiv.org/abs/2110.08207}.

\bibitem[Schr{\"o}der \& Schulz(2022)Schr{\"o}der and
  Schulz]{schroder2022monitoring}
Tim Schr{\"o}der and Michael Schulz.
\newblock Monitoring machine learning models: a categorization of challenges
  and methods.
\newblock \emph{Data Science and Management}, 5\penalty0 (3):\penalty0
  105--116, 2022.

\bibitem[Sculley et~al.(2015)Sculley, Holt, Golovin, Davydov, Phillips, Ebner,
  Chaudhary, Young, Crespo, and Dennison]{sculley2015hidden}
David Sculley, Gary Holt, Daniel Golovin, Eugene Davydov, Todd Phillips,
  Dietmar Ebner, Vinay Chaudhary, Michael Young, Jean-Francois Crespo, and Dan
  Dennison.
\newblock Hidden technical debt in machine learning systems.
\newblock \emph{Advances in neural information processing systems}, 28, 2015.

\bibitem[Serra et~al.(2018)Serra, Suris, Miron, and
  Karatzoglou]{serra2018overcoming}
Joan Serra, Didac Suris, Marius Miron, and Alexandros Karatzoglou.
\newblock Overcoming catastrophic forgetting with hard attention to the task.
\newblock In Jennifer Dy and Andreas Krause (eds.), \emph{Proceedings of the
  35th International Conference on Machine Learning}, volume~80 of
  \emph{Proceedings of Machine Learning Research}, pp.\  4548--4557. PMLR,
  10--15 Jul 2018.
\newblock URL \url{https://proceedings.mlr.press/v80/serra18a.html}.

\bibitem[Sharir et~al.(2020)Sharir, Peleg, and Shoham]{sharir2020cost}
Or~Sharir, Barak Peleg, and Yoav Shoham.
\newblock The cost of training nlp models: A concise overview.
\newblock \emph{arXiv preprint arXiv:2004.08900}, 2020.

\bibitem[Shin et~al.(2020)Shin, Razeghi, Logan~IV, Wallace, and
  Singh]{shin2020autoprompt}
Taylor Shin, Yasaman Razeghi, Robert~L. Logan~IV, Eric Wallace, and Sameer
  Singh.
\newblock {A}uto{P}rompt: {E}liciting {K}nowledge from {L}anguage {M}odels with
  {A}utomatically {G}enerated {P}rompts.
\newblock In \emph{Proceedings of the 2020 Conference on Empirical Methods in
  Natural Language Processing (EMNLP)}, pp.\  4222--4235, Online, November
  2020. Association for Computational Linguistics.
\newblock \doi{10.18653/v1/2020.emnlp-main.346}.
\newblock URL \url{https://aclanthology.org/2020.emnlp-main.346}.

\bibitem[Sinha et~al.(2018)Sinha, Chen, Badrinarayanan, and
  Rabinovich]{sinha2018gradient}
Ayan Sinha, Zhao Chen, Vijay Badrinarayanan, and Andrew Rabinovich.
\newblock Gradient adversarial training of neural networks.
\newblock \emph{CoRR}, abs/1806.08028, 2018.
\newblock URL \url{http://arxiv.org/abs/1806.08028}.

\bibitem[Standley et~al.(2020)Standley, Zamir, Chen, Guibas, Malik, and
  Savarese]{standley2020which}
Trevor Standley, Amir Zamir, Dawn Chen, Leonidas Guibas, Jitendra Malik, and
  Silvio Savarese.
\newblock Which tasks should be learned together in multi-task learning?
\newblock In Hal~Daumé III and Aarti Singh (eds.), \emph{Proceedings of the
  37th International Conference on Machine Learning}, volume 119 of
  \emph{Proceedings of Machine Learning Research}, pp.\  9120--9132. PMLR,
  13--18 Jul 2020.
\newblock URL \url{https://proceedings.mlr.press/v119/standley20a.html}.

\bibitem[Stickland \& Murray(2019)Stickland and Murray]{stickland2019bert}
Asa~Cooper Stickland and Iain Murray.
\newblock Bert and pals: Projected attention layers for efficient adaptation in
  multi-task learning.
\newblock In \emph{International Conference on Machine Learning}, pp.\
  5986--5995. PMLR, 2019.

\bibitem[Strubell et~al.(2020)Strubell, Ganesh, and
  McCallum]{strubell2020energy}
Emma Strubell, Ananya Ganesh, and Andrew McCallum.
\newblock Energy and policy considerations for modern deep learning research.
\newblock In \emph{Proceedings of the AAAI Conference on Artificial
  Intelligence}, volume~34, pp.\  13693--13696, 2020.

\bibitem[Sun(2020)]{sun2020optimization}
Ruo-Yu Sun.
\newblock Optimization for deep learning: An overview.
\newblock \emph{Journal of the Operations Research Society of China},
  8\penalty0 (2):\penalty0 249--294, 2020.

\bibitem[Sun et~al.(2020)Sun, Wang, Li, Feng, Tian, Wu, and Wang]{sun2020ernie}
Yu~Sun, Shuohuan Wang, Yukun Li, Shikun Feng, Hao Tian, Hua Wu, and Haifeng
  Wang.
\newblock Ernie 2.0: A continual pre-training framework for language
  understanding.
\newblock In \emph{Proceedings of the AAAI conference on artificial
  intelligence}, volume~34, pp.\  8968--8975, 2020.

\bibitem[Takeuchi \& Yamamoto(2020)Takeuchi and Yamamoto]{takeuchi2020business}
Hironori Takeuchi and Shuichiro Yamamoto.
\newblock Business analysis method for constructing business--ai alignment
  model.
\newblock \emph{Procedia Computer Science}, 176:\penalty0 1312--1321, 2020.

\bibitem[Tay et~al.(2020)Tay, Zhao, Bahri, Metzler, and Juan]{tay2020hypergrid}
Yi~Tay, Zhe Zhao, Dara Bahri, Donald Metzler, and Da{-}Cheng Juan.
\newblock Hypergrid: Efficient multi-task transformers with grid-wise
  decomposable hyper projections.
\newblock \emph{CoRR}, abs/2007.05891, 2020.
\newblock URL \url{https://arxiv.org/abs/2007.05891}.

\bibitem[Thung \& Wee(2018)Thung and Wee]{thung2018review}
Kim-Han Thung and Chong-Yaw Wee.
\newblock A brief review on multi-task learning.
\newblock \emph{Multimedia Tools and Applications}, 77\penalty0 (22):\penalty0
  29705--29725, 2018.

\bibitem[Upadhyay et~al.(2021)Upadhyay, Phlypo, Saini, and
  Liwicki]{upadhyay2021sharing}
Richa Upadhyay, Ronald Phlypo, Rajkumar Saini, and Marcus Liwicki.
\newblock Sharing to learn and learning to share-fitting together
  meta-learning, multi-task learning, and transfer learning: A meta review.
\newblock \emph{arXiv preprint arXiv:2111.12146}, 2021.

\bibitem[Vafaeikia et~al.(2020)Vafaeikia, Namdar, and
  Khalvati]{vafaeikia2020brief}
Partoo Vafaeikia, Khashayar Namdar, and Farzad Khalvati.
\newblock A brief review of deep multi-task learning and auxiliary task
  learning.
\newblock \emph{arXiv preprint arXiv:2007.01126}, 2020.

\bibitem[Vandenhende et~al.(2021)Vandenhende, Georgoulis, Van~Gansbeke,
  Proesmans, Dai, and Van~Gool]{vandenhende2021multi}
Simon Vandenhende, Stamatios Georgoulis, Wouter Van~Gansbeke, Marc Proesmans,
  Dengxin Dai, and Luc Van~Gool.
\newblock Multi-task learning for dense prediction tasks: A survey.
\newblock \emph{IEEE transactions on pattern analysis and machine
  intelligence}, 2021.

\bibitem[Vartak \& Madden(2018)Vartak and Madden]{vartak2018modeldb}
Manasi Vartak and Samuel Madden.
\newblock Modeldb: Opportunities and challenges in managing machine learning
  models.
\newblock \emph{IEEE Data Eng. Bull.}, 41\penalty0 (4):\penalty0 16--25, 2018.

\bibitem[Vaswani et~al.(2017)Vaswani, Shazeer, Parmar, Uszkoreit, Jones, Gomez,
  Kaiser, and Polosukhin]{vaswani2017attention}
Ashish Vaswani, Noam Shazeer, Niki Parmar, Jakob Uszkoreit, Llion Jones,
  Aidan~N Gomez, {\L}ukasz Kaiser, and Illia Polosukhin.
\newblock Attention is all you need.
\newblock \emph{Advances in neural information processing systems}, 30, 2017.

\bibitem[Von~Oswald et~al.(2019)Von~Oswald, Henning, Sacramento, and
  Grewe]{von2019continual}
Johannes Von~Oswald, Christian Henning, Jo{\~a}o Sacramento, and Benjamin~F
  Grewe.
\newblock Continual learning with hypernetworks.
\newblock \emph{arXiv preprint arXiv:1906.00695}, 2019.

\bibitem[Vu et~al.(2020)Vu, Wang, Munkhdalai, Sordoni, Trischler,
  Mattarella{-}Micke, Maji, and Iyyer]{vu2020exploring}
Tu~Vu, Tong Wang, Tsendsuren Munkhdalai, Alessandro Sordoni, Adam Trischler,
  Andrew Mattarella{-}Micke, Subhransu Maji, and Mohit Iyyer.
\newblock Exploring and predicting transferability across {NLP} tasks.
\newblock \emph{CoRR}, abs/2005.00770, 2020.
\newblock URL \url{https://arxiv.org/abs/2005.00770}.

\bibitem[Wang et~al.(2018)Wang, Singh, Michael, Hill, Levy, and
  Bowman]{wang2018glue}
Alex Wang, Amanpreet Singh, Julian Michael, Felix Hill, Omer Levy, and Samuel~R
  Bowman.
\newblock Glue: A multi-task benchmark and analysis platform for natural
  language understanding.
\newblock \emph{arXiv preprint arXiv:1804.07461}, 2018.

\bibitem[Wang et~al.(2019)Wang, Pruksachatkun, Nangia, Singh, Michael, Hill,
  Levy, and Bowman]{wang2019superglue}
Alex Wang, Yada Pruksachatkun, Nikita Nangia, Amanpreet Singh, Julian Michael,
  Felix Hill, Omer Levy, and Samuel Bowman.
\newblock Superglue: A stickier benchmark for general-purpose language
  understanding systems.
\newblock \emph{Advances in neural information processing systems}, 32, 2019.

\bibitem[Wang et~al.(2022)Wang, Li, Yan, Yan, Wang, Wu, and
  Xu]{wang2022instructionner}
Liwen Wang, Rumei Li, Yang Yan, Yuanmeng Yan, Sirui Wang, Wei Wu, and Weiran
  Xu.
\newblock Instructionner: A multi-task instruction-based generative framework
  for few-shot ner, 2022.
\newblock URL \url{https://arxiv.org/abs/2203.03903}.

\bibitem[Wang et~al.(2020)Wang, Khabsa, and Ma]{wang2020pretrain}
Sinong Wang, Madian Khabsa, and Hao Ma.
\newblock To pretrain or not to pretrain: Examining the benefits of pretraining
  on resource rich tasks.
\newblock \emph{arXiv preprint arXiv:2006.08671}, 2020.

\bibitem[Whang et~al.(2023)Whang, Roh, Song, and Lee]{whang2023data}
Steven~Euijong Whang, Yuji Roh, Hwanjun Song, and Jae-Gil Lee.
\newblock Data collection and quality challenges in deep learning: A
  data-centric ai perspective.
\newblock \emph{The VLDB Journal}, pp.\  1--23, 2023.

\bibitem[Worsham \& Kalita(2020)Worsham and Kalita]{worsham2020multi}
Joseph Worsham and Jugal Kalita.
\newblock Multi-task learning for natural language processing in the 2020s:
  where are we going?
\newblock \emph{Pattern Recognition Letters}, 136:\penalty0 120--126, 2020.

\bibitem[Wu \& Xie(2022)Wu and Xie]{wu2022survey}
Nan Wu and Yuan Xie.
\newblock A survey of machine learning for computer architecture and systems.
\newblock \emph{ACM Computing Surveys (CSUR)}, 55\penalty0 (3):\penalty0 1--39,
  2022.

\bibitem[Wu et~al.(2020)Wu, Zhang, and Ré]{wu2020understanding}
Sen Wu, Hongyang~R. Zhang, and Christopher Ré.
\newblock Understanding and improving information transfer in multi-task
  learning, 2020.
\newblock URL \url{https://arxiv.org/abs/2005.00944}.

\bibitem[Yang et~al.(2022)Yang, Brower-Sinning, Lewis, K{\"a}stner, and
  Wu]{yang2022capabilities}
Chenyang Yang, Rachel Brower-Sinning, Grace~A Lewis, Christian K{\"a}stner, and
  Tongshuang Wu.
\newblock Capabilities for better ml engineering.
\newblock \emph{arXiv preprint arXiv:2211.06409}, 2022.

\bibitem[Yang \& Hospedales(2016)Yang and Hospedales]{yongxin2016trace}
Yongxin Yang and Timothy~M. Hospedales.
\newblock Trace norm regularised deep multi-task learning.
\newblock \emph{CoRR}, abs/1606.04038, 2016.
\newblock URL \url{http://arxiv.org/abs/1606.04038}.

\bibitem[Zamir et~al.(2018)Zamir, Sax, Shen, Guibas, Malik, and
  Savarese]{Zamir_2018_CVPR}
Amir~R. Zamir, Alexander Sax, William Shen, Leonidas~J. Guibas, Jitendra Malik,
  and Silvio Savarese.
\newblock Taskonomy: Disentangling task transfer learning.
\newblock In \emph{Proceedings of the IEEE Conference on Computer Vision and
  Pattern Recognition (CVPR)}, June 2018.

\bibitem[Zhai et~al.(2019)Zhai, Wu, Tzeng, Park, and
  Rosenberg]{zhai2019learning}
Andrew Zhai, Hao-Yu Wu, Eric Tzeng, Dong~Huk Park, and Charles Rosenberg.
\newblock Learning a unified embedding for visual search at pinterest.
\newblock In \emph{Proceedings of the 25th ACM SIGKDD International Conference
  on Knowledge Discovery \& Data Mining}, pp.\  2412--2420, 2019.

\bibitem[Zhang \& Yang(2017)Zhang and Yang]{zhang2017survey}
Yu~Zhang and Qiang Yang.
\newblock A survey on multi-task learning.
\newblock \emph{arXiv preprint arXiv:1707.08114}, 2017.

\bibitem[Zhang \& Yang(2018)Zhang and Yang]{zhang2018overview}
Yu~Zhang and Qiang Yang.
\newblock An overview of multi-task learning.
\newblock \emph{National Science Review}, 5\penalty0 (1):\penalty0 30--43,
  2018.

\bibitem[Zhang et~al.(2023)Zhang, Yu, Yu, Guo, and Jiang]{zhang2023survey}
Zhihan Zhang, Wenhao Yu, Mengxia Yu, Zhichun Guo, and Meng Jiang.
\newblock A survey of multi-task learning in natural language processing:
  Regarding task relatedness and training methods.
\newblock In \emph{Proceedings of the 17th Conference of the European Chapter
  of the Association for Computational Linguistics}, pp.\  943--956, Dubrovnik,
  Croatia, May 2023. Association for Computational Linguistics.
\newblock URL \url{https://aclanthology.org/2023.eacl-main.66}.

\bibitem[Zheng et~al.(2019)Zheng, Deng, Sun, Jiang, Guo, Yu, Huang, and
  Ji]{Zheng_2019_CVPR}
Feng Zheng, Cheng Deng, Xing Sun, Xinyang Jiang, Xiaowei Guo, Zongqiao Yu,
  Feiyue Huang, and Rongrong Ji.
\newblock Pyramidal person re-identification via multi-loss dynamic training.
\newblock In \emph{Proceedings of the IEEE/CVF Conference on Computer Vision
  and Pattern Recognition (CVPR)}, June 2019.

\bibitem[Zhou(2019)]{zhou2019overview}
Wenxuan Zhou.
\newblock An overview of models and methods for multi-task learning, Oct 2019.
\newblock URL
  \url{https://shanzhenren.github.io/csci-699-replnlp-2019fall/lectures/W6-L1-Multi_Task_Learning.pdf}.

\end{thebibliography}
\bibliographystyle{paper}

\appendix

\section{Additional Details for Related Surveys}

\subsection{Multi-Task Learning Surveys}
\label{subsection:appendix-mtl-surveys}

\begin{table}[t]
\caption{Discussed aspects per MTL survey. Aspects are indicated in \textbf{bold}.}
\label{table:mtl-surveys-overview-apendix}
\begin{center}
\begin{small}

    \begin{tabular}{lllllllllllllll}
        \hline
        \multicolumn{1}{c|}{Year}
        & \multicolumn{14}{c}{MTL Survey} \\
        \hline
        \multicolumn{1}{c|}{$2017$}
        & \multicolumn{14}{c}{
            $1$ - \citep{ruder2017overview}
            $2$ - \citep{zhang2017survey}
        } \\
        \multicolumn{1}{c|}{$2018$}
        & \multicolumn{14}{c}{
            $3$ - \citep{zhang2018overview}
            $4$ - \citep{thung2018review}     
        } \\
        \multicolumn{1}{c|}{$2019$}
        & \multicolumn{14}{c}{
            $5$ - \citep{zhou2019overview}  
        } \\
        \multicolumn{1}{c|}{$2020$}
        & \multicolumn{14}{c}{
            $6$ - \citep{vafaeikia2020brief}
            $7$ - \citep{worsham2020multi}
            $8$ - \citep{crawshaw2020multi}  
        } \\
        \multicolumn{1}{c|}{$2021$}
        & \multicolumn{14}{c}{
            $9$ - \citep{vandenhende2021multi}
            $10$ - \citep{chen2021multi}
            $11$ - \citep{upadhyay2021sharing}   
        } \\
        \multicolumn{1}{c|}{$2022$}
        & \multicolumn{14}{c}{
            $12$ - \citep{samant2022framework}
            $13$ - \citep{abhadiomhen2022supervised}
        } \\
        \multicolumn{1}{c|}{$2023$}
        & \multicolumn{14}{c}{
            $14$ - \citep{zhang2023survey}
        } \\
        \hline
        \multicolumn{1}{c}{Aspect \textbackslash Survey}
        &\multicolumn{1}{c}{$1$}
        &\multicolumn{1}{c}{$2$}
        &\multicolumn{1}{c}{$3$}
        &\multicolumn{1}{c}{$4$}
        &\multicolumn{1}{c}{$5$}
        &\multicolumn{1}{c}{$6$}
        &\multicolumn{1}{c}{$7$}
        &\multicolumn{1}{c}{$8$}
        &\multicolumn{1}{c}{$9$}
        &\multicolumn{1}{c}{$10$}
        &\multicolumn{1}{c}{$11$}
        &\multicolumn{1}{c}{$12$}
        &\multicolumn{1}{c}{$13$}
        &\multicolumn{1}{c}{$14$}
        \\ \hline
        \multicolumn{1}{c}{\textbf{Computational Model}}
        &\multicolumn{14}{c}{}
        \\ \hline
        \multicolumn{1}{c|}{Traditional ML}
        &\multicolumn{1}{c}{\checkmark}
        &\multicolumn{1}{c}{\checkmark}
        &\multicolumn{1}{c}{\checkmark}
        &\multicolumn{1}{c}{\checkmark}
        &\multicolumn{1}{c}{}
        &\multicolumn{1}{c}{}
        &\multicolumn{1}{c}{}
        &\multicolumn{1}{c}{}
        &\multicolumn{1}{c}{}
        &\multicolumn{1}{c}{}
        &\multicolumn{1}{c}{}
        &\multicolumn{1}{c}{}
        &\multicolumn{1}{c}{\checkmark}
        &\multicolumn{1}{c}{}
        \\
        \multicolumn{1}{c|}{Deep Learning}
        &\multicolumn{1}{c}{\checkmark}
        &\multicolumn{1}{c}{\checkmark}
        &\multicolumn{1}{c}{\checkmark}
        &\multicolumn{1}{c}{\checkmark}
        &\multicolumn{1}{c}{\checkmark}
        &\multicolumn{1}{c}{\checkmark}
        &\multicolumn{1}{c}{\checkmark}
        &\multicolumn{1}{c}{\checkmark}
        &\multicolumn{1}{c}{\checkmark}
        &\multicolumn{1}{c}{\checkmark}
        &\multicolumn{1}{c}{\checkmark}
        &\multicolumn{1}{c}{\checkmark}
        &\multicolumn{1}{c}{}
        &\multicolumn{1}{c}{\checkmark}
        \\ \hline
        \multicolumn{1}{c}{\textbf{Learning Type}}
        &\multicolumn{14}{c}{}
        \\ \hline
        \multicolumn{1}{c|}{Joint Learning}
        &\multicolumn{1}{c}{\checkmark}
        &\multicolumn{1}{c}{\checkmark}
        &\multicolumn{1}{c}{\checkmark}
        &\multicolumn{1}{c}{\checkmark}
        &\multicolumn{1}{c}{\checkmark}
        &\multicolumn{1}{c}{\checkmark}
        &\multicolumn{1}{c}{\checkmark}
        &\multicolumn{1}{c}{\checkmark}
        &\multicolumn{1}{c}{\checkmark}
        &\multicolumn{1}{c}{\checkmark}
        &\multicolumn{1}{c}{\checkmark}
        &\multicolumn{1}{c}{\checkmark}
        &\multicolumn{1}{c}{\checkmark}
        &\multicolumn{1}{c}{\checkmark}
        \\
        \multicolumn{1}{c|}{Auxiliary Learning}
        &\multicolumn{1}{c}{\checkmark}
        &\multicolumn{1}{c}{\checkmark}
        &\multicolumn{1}{c}{\checkmark}
        &\multicolumn{1}{c}{\checkmark}
        &\multicolumn{1}{c}{\checkmark}
        &\multicolumn{1}{c}{\checkmark}
        &\multicolumn{1}{c}{\checkmark}
        &\multicolumn{1}{c}{}
        &\multicolumn{1}{c}{}
        &\multicolumn{1}{c}{\checkmark}
        &\multicolumn{1}{c}{}
        &\multicolumn{1}{c}{\checkmark}
        &\multicolumn{1}{c}{}
        &\multicolumn{1}{c}{\checkmark}
        \\ \hline
        \multicolumn{1}{c}{\textbf{Architectures}}
        &\multicolumn{14}{c}{}
        \\ \hline
        \multicolumn{1}{c|}{Taxonomy}
        &\multicolumn{1}{c}{\checkmark}
        &\multicolumn{1}{c}{\checkmark}
        &\multicolumn{1}{c}{\checkmark}
        &\multicolumn{1}{c}{\checkmark}
        &\multicolumn{1}{c}{\checkmark}
        &\multicolumn{1}{c}{\checkmark}
        &\multicolumn{1}{c}{\checkmark}
        &\multicolumn{1}{c}{\checkmark}
        &\multicolumn{1}{c}{\checkmark}
        &\multicolumn{1}{c}{\checkmark}
        &\multicolumn{1}{c}{\checkmark}
        &\multicolumn{1}{c}{\checkmark}
        &\multicolumn{1}{c}{\checkmark}
        &\multicolumn{1}{c}{\checkmark}
        \\
        \multicolumn{1}{c|}{Learning to Share}
        &\multicolumn{1}{c}{\checkmark}
        &\multicolumn{1}{c}{}
        &\multicolumn{1}{c}{}
        &\multicolumn{1}{c}{\checkmark}
        &\multicolumn{1}{c}{\checkmark}
        &\multicolumn{1}{c}{\checkmark}
        &\multicolumn{1}{c}{}
        &\multicolumn{1}{c}{\checkmark}
        &\multicolumn{1}{c}{\checkmark}
        &\multicolumn{1}{c}{\checkmark}
        &\multicolumn{1}{c}{}
        &\multicolumn{1}{c}{}
        &\multicolumn{1}{c}{}
        &\multicolumn{1}{c}{\checkmark}
        \\
        \multicolumn{1}{c|}{Universal Models}
        &\multicolumn{1}{c}{}
        &\multicolumn{1}{c}{}
        &\multicolumn{1}{c}{}
        &\multicolumn{1}{c}{}
        &\multicolumn{1}{c}{\checkmark}
        &\multicolumn{1}{c}{}
        &\multicolumn{1}{c}{}
        &\multicolumn{1}{c}{\checkmark}
        &\multicolumn{1}{c}{\checkmark}
        &\multicolumn{1}{c}{\checkmark}
        &\multicolumn{1}{c}{}
        &\multicolumn{1}{c}{}
        &\multicolumn{1}{c}{}
        &\multicolumn{1}{c}{\checkmark}
        \\ \hline
        \multicolumn{1}{c}{\textbf{Optimization}}
        &\multicolumn{14}{c}{}
        \\ \hline
        \multicolumn{1}{c|}{Loss Weighting}
        &\multicolumn{1}{c}{\checkmark}
        &\multicolumn{1}{c}{\checkmark}
        &\multicolumn{1}{c}{\checkmark}
        &\multicolumn{1}{c}{}
        &\multicolumn{1}{c}{\checkmark}
        &\multicolumn{1}{c}{\checkmark}
        &\multicolumn{1}{c}{\checkmark}
        &\multicolumn{1}{c}{\checkmark}
        &\multicolumn{1}{c}{\checkmark}
        &\multicolumn{1}{c}{\checkmark}
        &\multicolumn{1}{c}{}
        &\multicolumn{1}{c}{\checkmark}
        &\multicolumn{1}{c}{}
        &\multicolumn{1}{c}{}
        \\
        \multicolumn{1}{c|}{Regularization}
        &\multicolumn{1}{c}{\checkmark}
        &\multicolumn{1}{c}{\checkmark}
        &\multicolumn{1}{c}{\checkmark}
        &\multicolumn{1}{c}{\checkmark}
        &\multicolumn{1}{c}{\checkmark}
        &\multicolumn{1}{c}{}
        &\multicolumn{1}{c}{}
        &\multicolumn{1}{c}{\checkmark}
        &\multicolumn{1}{c}{}
        &\multicolumn{1}{c}{\checkmark}
        &\multicolumn{1}{c}{}
        &\multicolumn{1}{c}{}
        &\multicolumn{1}{c}{\checkmark}
        &\multicolumn{1}{c}{}
        \\
        \multicolumn{1}{c|}{Task Scheduling}
        &\multicolumn{1}{c}{}
        &\multicolumn{1}{c}{}
        &\multicolumn{1}{c}{}
        &\multicolumn{1}{c}{}
        &\multicolumn{1}{c}{\checkmark}
        &\multicolumn{1}{c}{\checkmark}
        &\multicolumn{1}{c}{\checkmark}
        &\multicolumn{1}{c}{\checkmark}
        &\multicolumn{1}{c}{\checkmark}
        &\multicolumn{1}{c}{\checkmark}
        &\multicolumn{1}{c}{}
        &\multicolumn{1}{c}{}
        &\multicolumn{1}{c}{}
        &\multicolumn{1}{c}{}
        \\
        \multicolumn{1}{c|}{Gradient Modulation}
        &\multicolumn{1}{c}{\checkmark}
        &\multicolumn{1}{c}{}
        &\multicolumn{1}{c}{}
        &\multicolumn{1}{c}{}
        &\multicolumn{1}{c}{}
        &\multicolumn{1}{c}{\checkmark}
        &\multicolumn{1}{c}{}
        &\multicolumn{1}{c}{\checkmark}
        &\multicolumn{1}{c}{\checkmark}
        &\multicolumn{1}{c}{\checkmark}
        &\multicolumn{1}{c}{}
        &\multicolumn{1}{c}{}
        &\multicolumn{1}{c}{}
        &\multicolumn{1}{c}{}
        \\
        \multicolumn{1}{c|}{Knowledge Distillation}
        &\multicolumn{1}{c}{}
        &\multicolumn{1}{c}{}
        &\multicolumn{1}{c}{}
        &\multicolumn{1}{c}{}
        &\multicolumn{1}{c}{}
        &\multicolumn{1}{c}{}
        &\multicolumn{1}{c}{\checkmark}
        &\multicolumn{1}{c}{\checkmark}
        &\multicolumn{1}{c}{}
        &\multicolumn{1}{c}{\checkmark}
        &\multicolumn{1}{c}{}
        &\multicolumn{1}{c}{\checkmark}
        &\multicolumn{1}{c}{}
        &\multicolumn{1}{c}{}
        \\
        \multicolumn{1}{c|}{Multi-Objective Optimization}
        &\multicolumn{1}{c}{}
        &\multicolumn{1}{c}{}
        &\multicolumn{1}{c}{}
        &\multicolumn{1}{c}{}
        &\multicolumn{1}{c}{}
        &\multicolumn{1}{c}{}
        &\multicolumn{1}{c}{}
        &\multicolumn{1}{c}{\checkmark}
        &\multicolumn{1}{c}{\checkmark}
        &\multicolumn{1}{c}{\checkmark}
        &\multicolumn{1}{c}{}
        &\multicolumn{1}{c}{}
        &\multicolumn{1}{c}{}
        &\multicolumn{1}{c}{}
        \\ \hline
        \multicolumn{1}{c}{\textbf{Task Relationship Learning}}
        &\multicolumn{14}{c}{}
        \\ \hline
        \multicolumn{1}{c|}{Task Grouping}
        &\multicolumn{1}{c}{\checkmark}
        &\multicolumn{1}{c}{\checkmark}
        &\multicolumn{1}{c}{\checkmark}
        &\multicolumn{1}{c}{\checkmark}
        &\multicolumn{1}{c}{}
        &\multicolumn{1}{c}{}
        &\multicolumn{1}{c}{\checkmark}
        &\multicolumn{1}{c}{\checkmark}
        &\multicolumn{1}{c}{\checkmark}
        &\multicolumn{1}{c}{}
        &\multicolumn{1}{c}{}
        &\multicolumn{1}{c}{}
        &\multicolumn{1}{c}{\checkmark}
        &\multicolumn{1}{c}{\checkmark}
        \\
        \multicolumn{1}{c|}{Relationships Transfer}
        &\multicolumn{1}{c}{\checkmark}
        &\multicolumn{1}{c}{}
        &\multicolumn{1}{c}{}
        &\multicolumn{1}{c}{\checkmark}
        &\multicolumn{1}{c}{}
        &\multicolumn{1}{c}{}
        &\multicolumn{1}{c}{\checkmark}
        &\multicolumn{1}{c}{\checkmark}
        &\multicolumn{1}{c}{\checkmark}
        &\multicolumn{1}{c}{}
        &\multicolumn{1}{c}{}
        &\multicolumn{1}{c}{}
        &\multicolumn{1}{c}{}
        &\multicolumn{1}{c}{}
        \\
        \multicolumn{1}{c|}{Task Embeddings}
        &\multicolumn{1}{c}{}
        &\multicolumn{1}{c}{}
        &\multicolumn{1}{c}{}
        &\multicolumn{1}{c}{}
        &\multicolumn{1}{c}{}
        &\multicolumn{1}{c}{}
        &\multicolumn{1}{c}{\checkmark}
        &\multicolumn{1}{c}{\checkmark}
        &\multicolumn{1}{c}{}
        &\multicolumn{1}{c}{}
        &\multicolumn{1}{c}{}
        &\multicolumn{1}{c}{}
        &\multicolumn{1}{c}{}
        &\multicolumn{1}{c}{}
        \\ \hline
        \multicolumn{1}{c}{\textbf{Supervision Level}}
        &\multicolumn{14}{c}{}
        \\ \hline
        \multicolumn{1}{c|}{Supervised Learning}
        &\multicolumn{1}{c}{\checkmark}
        &\multicolumn{1}{c}{\checkmark}
        &\multicolumn{1}{c}{\checkmark}
        &\multicolumn{1}{c}{\checkmark}
        &\multicolumn{1}{c}{\checkmark}
        &\multicolumn{1}{c}{\checkmark}
        &\multicolumn{1}{c}{\checkmark}
        &\multicolumn{1}{c}{\checkmark}
        &\multicolumn{1}{c}{\checkmark}
        &\multicolumn{1}{c}{\checkmark}
        &\multicolumn{1}{c}{\checkmark}
        &\multicolumn{1}{c}{\checkmark}
        &\multicolumn{1}{c}{\checkmark}
        &\multicolumn{1}{c}{\checkmark}
        \\
        \multicolumn{1}{c|}{Semi-supervised Learning}
        &\multicolumn{1}{c}{}
        &\multicolumn{1}{c}{\checkmark}
        &\multicolumn{1}{c}{\checkmark}
        &\multicolumn{1}{c}{}
        &\multicolumn{1}{c}{}
        &\multicolumn{1}{c}{}
        &\multicolumn{1}{c}{}
        &\multicolumn{1}{c}{}
        &\multicolumn{1}{c}{}
        &\multicolumn{1}{c}{\checkmark}
        &\multicolumn{1}{c}{}
        &\multicolumn{1}{c}{\checkmark}
        &\multicolumn{1}{c}{}
        &\multicolumn{1}{c}{\checkmark}
        \\
        \multicolumn{1}{c|}{Self-supervised Learning}
        &\multicolumn{1}{c}{}
        &\multicolumn{1}{c}{\checkmark}
        &\multicolumn{1}{c}{\checkmark}
        &\multicolumn{1}{c}{}
        &\multicolumn{1}{c}{}
        &\multicolumn{1}{c}{}
        &\multicolumn{1}{c}{\checkmark}
        &\multicolumn{1}{c}{}
        &\multicolumn{1}{c}{}
        &\multicolumn{1}{c}{\checkmark}
        &\multicolumn{1}{c}{}
        &\multicolumn{1}{c}{\checkmark}
        &\multicolumn{1}{c}{}
        &\multicolumn{1}{c}{\checkmark}
        \\ \hline
        \multicolumn{1}{c}{\textbf{Connection to Learning Paradigm}}
        &\multicolumn{14}{c}{}
        \\ \hline
        \multicolumn{1}{c|}{Reinforcement Learning}
        &\multicolumn{1}{c}{}
        &\multicolumn{1}{c}{\checkmark}
        &\multicolumn{1}{c}{\checkmark}
        &\multicolumn{1}{c}{}
        &\multicolumn{1}{c}{\checkmark}
        &\multicolumn{1}{c}{\checkmark}
        &\multicolumn{1}{c}{}
        &\multicolumn{1}{c}{\checkmark}
        &\multicolumn{1}{c}{}
        &\multicolumn{1}{c}{}
        &\multicolumn{1}{c}{}
        &\multicolumn{1}{c}{}
        &\multicolumn{1}{c}{}
        &\multicolumn{1}{c}{\checkmark}
        \\
        \multicolumn{1}{c|}{Transfer Learning}
        &\multicolumn{1}{c}{}
        &\multicolumn{1}{c}{\checkmark}
        &\multicolumn{1}{c}{\checkmark}
        &\multicolumn{1}{c}{\checkmark}
        &\multicolumn{1}{c}{}
        &\multicolumn{1}{c}{}
        &\multicolumn{1}{c}{}
        &\multicolumn{1}{c}{}
        &\multicolumn{1}{c}{}
        &\multicolumn{1}{c}{}
        &\multicolumn{1}{c}{}
        &\multicolumn{1}{c}{}
        &\multicolumn{1}{c}{}
        &\multicolumn{1}{c}{}
        \\
        \multicolumn{1}{c|}{Domain Adaptation}
        &\multicolumn{1}{c}{\checkmark}
        &\multicolumn{1}{c}{}
        &\multicolumn{1}{c}{}
        &\multicolumn{1}{c}{}
        &\multicolumn{1}{c}{}
        &\multicolumn{1}{c}{}
        &\multicolumn{1}{c}{\checkmark}
        &\multicolumn{1}{c}{}
        &\multicolumn{1}{c}{}
        &\multicolumn{1}{c}{}
        &\multicolumn{1}{c}{\checkmark}
        &\multicolumn{1}{c}{}
        &\multicolumn{1}{c}{}
        &\multicolumn{1}{c}{}
        \\
        \multicolumn{1}{c|}{Meta-Learning}
        &\multicolumn{1}{c}{}
        &\multicolumn{1}{c}{}
        &\multicolumn{1}{c}{}
        &\multicolumn{1}{c}{}
        &\multicolumn{1}{c}{}
        &\multicolumn{1}{c}{}
        &\multicolumn{1}{c}{}
        &\multicolumn{1}{c}{\checkmark}
        &\multicolumn{1}{c}{}
        &\multicolumn{1}{c}{}
        &\multicolumn{1}{c}{\checkmark}
        &\multicolumn{1}{c}{}
        &\multicolumn{1}{c}{}
        &\multicolumn{1}{c}{\checkmark}
        \\
        \multicolumn{1}{c|}{Active Learning}
        &\multicolumn{1}{c}{}
        &\multicolumn{1}{c}{\checkmark}
        &\multicolumn{1}{c}{\checkmark}
        &\multicolumn{1}{c}{}
        &\multicolumn{1}{c}{}
        &\multicolumn{1}{c}{}
        &\multicolumn{1}{c}{}
        &\multicolumn{1}{c}{}
        &\multicolumn{1}{c}{}
        &\multicolumn{1}{c}{}
        &\multicolumn{1}{c}{}
        &\multicolumn{1}{c}{}
        &\multicolumn{1}{c}{}
        &\multicolumn{1}{c}{}
        \\
        \multicolumn{1}{c|}{Online Learning}
        &\multicolumn{1}{c}{\checkmark}
        &\multicolumn{1}{c}{\checkmark}
        &\multicolumn{1}{c}{\checkmark}
        &\multicolumn{1}{c}{}
        &\multicolumn{1}{c}{}
        &\multicolumn{1}{c}{}
        &\multicolumn{1}{c}{}
        &\multicolumn{1}{c}{}
        &\multicolumn{1}{c}{}
        &\multicolumn{1}{c}{}
        &\multicolumn{1}{c}{}
        &\multicolumn{1}{c}{}
        &\multicolumn{1}{c}{}
        &\multicolumn{1}{c}{}
        \\
        \multicolumn{1}{c|}{Continual Learning}
        &\multicolumn{1}{c}{}
        &\multicolumn{1}{c}{}
        &\multicolumn{1}{c}{}
        &\multicolumn{1}{c}{}
        &\multicolumn{1}{c}{}
        &\multicolumn{1}{c}{}
        &\multicolumn{1}{c}{}
        &\multicolumn{1}{c}{}
        &\multicolumn{1}{c}{}
        &\multicolumn{1}{c}{}
        &\multicolumn{1}{c}{}
        &\multicolumn{1}{c}{}
        &\multicolumn{1}{c}{}
        &\multicolumn{1}{c}{}
        \\ \hline
        \multicolumn{1}{c}{\textbf{Benchmarks}}
        &\multicolumn{14}{c}{}
        \\ \hline
        \multicolumn{1}{c|}{Benchmark Overview}
        &\multicolumn{1}{c}{}
        &\multicolumn{1}{c}{}
        &\multicolumn{1}{c}{\checkmark}
        &\multicolumn{1}{c}{}
        &\multicolumn{1}{c}{}
        &\multicolumn{1}{c}{}
        &\multicolumn{1}{c}{\checkmark}
        &\multicolumn{1}{c}{\checkmark}
        &\multicolumn{1}{c}{\checkmark}
        &\multicolumn{1}{c}{\checkmark}
        &\multicolumn{1}{c}{}
        &\multicolumn{1}{c}{\checkmark}
        &\multicolumn{1}{c}{\checkmark}
        &\multicolumn{1}{c}{}
        \\
        \multicolumn{1}{c|}{Model Comparison}
        &\multicolumn{1}{c}{}
        &\multicolumn{1}{c}{}
        &\multicolumn{1}{c}{\checkmark}
        &\multicolumn{1}{c}{}
        &\multicolumn{1}{c}{}
        &\multicolumn{1}{c}{}
        &\multicolumn{1}{c}{\checkmark}
        &\multicolumn{1}{c}{\checkmark}
        &\multicolumn{1}{c}{\checkmark}
        &\multicolumn{1}{c}{\checkmark}
        &\multicolumn{1}{c}{}
        &\multicolumn{1}{c}{\checkmark}
        &\multicolumn{1}{c}{\checkmark}
        &\multicolumn{1}{c}{}
        \\ \hline
        \multicolumn{1}{c}{\textbf{Application Domain}}
        &\multicolumn{14}{c}{}
        \\ \hline
        \multicolumn{1}{c|}{Natural Language Processing}
        &\multicolumn{1}{c}{\checkmark}
        &\multicolumn{1}{c}{\checkmark}
        &\multicolumn{1}{c}{\checkmark}
        &\multicolumn{1}{c}{\checkmark}
        &\multicolumn{1}{c}{\checkmark}
        &\multicolumn{1}{c}{}
        &\multicolumn{1}{c}{\checkmark}
        &\multicolumn{1}{c}{\checkmark}
        &\multicolumn{1}{c}{}
        &\multicolumn{1}{c}{\checkmark}
        &\multicolumn{1}{c}{\checkmark}
        &\multicolumn{1}{c}{\checkmark}
        &\multicolumn{1}{c}{}
        &\multicolumn{1}{c}{\checkmark}
        \\
        \multicolumn{1}{c|}{Computer Vision}
        &\multicolumn{1}{c}{\checkmark}
        &\multicolumn{1}{c}{\checkmark}
        &\multicolumn{1}{c}{\checkmark}
        &\multicolumn{1}{c}{\checkmark}
        &\multicolumn{1}{c}{\checkmark}
        &\multicolumn{1}{c}{\checkmark}
        &\multicolumn{1}{c}{}
        &\multicolumn{1}{c}{\checkmark}
        &\multicolumn{1}{c}{\checkmark}
        &\multicolumn{1}{c}{}
        &\multicolumn{1}{c}{\checkmark}
        &\multicolumn{1}{c}{}
        &\multicolumn{1}{c}{}
        &\multicolumn{1}{c}{}
        \\
        \multicolumn{1}{c|}{Healthcare}
        &\multicolumn{1}{c}{}
        &\multicolumn{1}{c}{\checkmark}
        &\multicolumn{1}{c}{\checkmark}
        &\multicolumn{1}{c}{\checkmark}
        &\multicolumn{1}{c}{}
        &\multicolumn{1}{c}{}
        &\multicolumn{1}{c}{}
        &\multicolumn{1}{c}{}
        &\multicolumn{1}{c}{}
        &\multicolumn{1}{c}{}
        &\multicolumn{1}{c}{}
        &\multicolumn{1}{c}{}
        &\multicolumn{1}{c}{}
        &\multicolumn{1}{c}{}
        \\
        \multicolumn{1}{c|}{Bioinformatics}
        &\multicolumn{1}{c}{\checkmark}
        &\multicolumn{1}{c}{\checkmark}
        &\multicolumn{1}{c}{\checkmark}
        &\multicolumn{1}{c}{\checkmark}
        &\multicolumn{1}{c}{}
        &\multicolumn{1}{c}{}
        &\multicolumn{1}{c}{}
        &\multicolumn{1}{c}{}
        &\multicolumn{1}{c}{}
        &\multicolumn{1}{c}{}
        &\multicolumn{1}{c}{\checkmark}
        &\multicolumn{1}{c}{}
        &\multicolumn{1}{c}{}
        &\multicolumn{1}{c}{}
        \\
        \multicolumn{1}{c|}{Other}
        &\multicolumn{1}{c}{\checkmark}
        &\multicolumn{1}{c}{\checkmark}
        &\multicolumn{1}{c}{\checkmark}
        &\multicolumn{1}{c}{}
        &\multicolumn{1}{c}{}
        &\multicolumn{1}{c}{}
        &\multicolumn{1}{c}{}
        &\multicolumn{1}{c}{\checkmark}
        &\multicolumn{1}{c}{}
        &\multicolumn{1}{c}{}
        &\multicolumn{1}{c}{}
        &\multicolumn{1}{c}{}
        &\multicolumn{1}{c}{\checkmark}
        &\multicolumn{1}{c}{}
        \\ \hline
    \end{tabular}

\end{small}
\end{center}
\end{table}

The idea of MTL has been explored in many research studies. In this section, we provide an overview of related work in MTL surveys and address different MTL aspects that were discussed. In Table~\ref{table:mtl-surveys-overview-apendix}, we list certain aspects and indicate the survey in which they were discussed. In the rest of the section, we just mention related surveys and go over individual MTL aspects (shown in \textbf{bold}) in more detail.

Many \textbf{application domains} were studied in previous work, ranging from surveys covering multiple domains  \citep{ruder2017overview,zhang2017survey,zhang2018overview,thung2018review,vafaeikia2020brief,crawshaw2020multi,upadhyay2021sharing,abhadiomhen2022supervised}, to those dedicated to a specific domain, such as computer vision \citep{vandenhende2021multi} or natural language processing \citep{zhou2019overview,worsham2020multi,chen2021multi,samant2022framework,zhang2023survey}.
Both traditional ML and deep learning \textbf{computational models} were studied. The traditional ML was discussed primarily in older studies, while deep learning was represented in all but one study.
Furthermore, most prior works provided an overview of the \textbf{benchmarks} for the specific domain \citep{mccann2018natural, wang2019superglue}, and compare models on them.

\textbf{Learning types.} Two learning types were mainly discussed. First, joint learning was used in an MTL setting where all the tasks are equally important \citep{Kendall2018CVPR, liu2019multitask}. Here, the goal is to achieve an on-par performance compared to their single-task learning (STL) counterparts. Second, auxiliary learning was used when there was a single main task or a set of them, while auxiliary tasks are only used to improve the performance of the main tasks \citep{González_Garduño_Søgaard_2018, wang2022instructionner}. Although the difference between the two learning types can be made, sometimes auxiliary learning was not differentiated from joint learning.

\textbf{Architectures}. MTL model architectures were among the most discussed aspects of MTL and one of the first aspects tackled in previous work. 
Distinguishing between hard and soft parameter sharing \citep{ruder2017overview} is the most used architecture taxonomy. 
In recent surveys, the method of sharing parameters between tasks has improved, leading to the refinement of taxonomies to categorize architectures more precisely \citep{crawshaw2020multi, chen2021multi}. 
Next, some MTL architectures were categorized as learning-to-share approaches \citep{ruder2017sluice}. 
Those works argue that it is better to learn parameter sharing in architectures for MTL rather than hand-design where sharing happens, as it offers a more adaptive solution to accommodate task similarities at different parts of the network.
Additionally, some MTL architectures were categorized as universal models which can handle multiple modalities, domains, and tasks with a single model \citep{kaiser2017one, subhojeet2019omninet}.

\textbf{Optimization}. Optimization techniques for MTL architectures were also discussed in detail.
To start with, the most common approach to mitigating MTL challenges is loss weighting. Computing weights of task-specific losses is crucial, as it helps optimize the loss function and considers the relative importance of each task.
There are various approaches to computing the loss weights dynamically, including weighting by uncertainty \citep{Kendall2018CVPR}, learning speed \citep{Liu_Liang_Gitter_2019, Zheng_2019_CVPR}, or performance \citep{Guo_2018_ECCV, Jean2019adaptive}, among others.
Next, and closely related to weighting task losses, is a task scheduling problem that involves choosing tasks to train on at each step. Many techniques were used, from simple ones that employ uniform or proportional task sampling, to the more complicated ones, such as annealed sampling \citep{stickland2019bert} or approaches based on active learning \citep{pilault2020conditionally}.
Regularization approaches were also analyzed. Methods include (1) minimizing the L2 norm between the parameters of soft-parameter sharing models \citep{yongxin2016trace}, (2) placing prior distributions on the network parameters \citep{long2017learning}, (3) introducing an auto-encoder term to the objective function \citep{lee2018deep}, (4) MTL variant of dropout \citep{Pascal_Michiardi_Bost_Huet_Zuluaga_2021}, and others.
To continue, gradient modulation techniques were employed to mitigate the problem of negative transfer by manipulating gradients of contradictory tasks, either through adversarial training \citep{sinha2018gradient} or by replacing the gradient by its modified version \citep{lopez_paz2017gradient}.
Another approach for optimizing MTL models is by applying knowledge distillation \citep{clark2019bam}.
Finally, multi-objective optimization has been applied to the MTL setting to obtain a set of Pareto optimal solutions on the Pareto frontier, providing greater flexibility in balancing trade-offs between tasks \citep{lin2019pareto}.

\textbf{Task relationship learning.} The approach in MTL that focuses on learning the explicit representation of tasks or relationships between them is task relationship learning. 
This approach consists of three main categories of methods. First, task grouping aims to divide a set of tasks into groups in order to maximize knowledge sharing during joint training \citep{standley2020which}. Second, transfer relationship learning involves methods that determine when transferring knowledge from one task to another will be beneficial for joint learning \citep{Zamir_2018_CVPR, guo2019autosem}. Finally, task embedding methods aim to learn an embedding space for the tasks themselves \citep{vu2020exploring}.

As for the \textbf{supervision level}, most studies focused on supervised approaches. Some studies analyzed semi-supervised approaches that incorporate self-supervised objectives, such as MLM. However, self-supervised approaches are mainly discussed in the context of pre-training. 

Most studies had made \textbf{connections to other learning paradigms}, including reinforcement learning, transfer learning with a special emphasis on domain adaptation, meta-learning, and active and online learning.
However, only \citet{ruder2017overview,zhang2017survey,zhang2018overview} explored the relationship between MTL and online learning in the context of traditional ML.
To the best of our knowledge, there had been no prior work systematically investigating connections between MTL and CL. We believe that a connection between MTL and CL represents a promising research direction, as we will motivate the importance of this connection in Section~\ref{section:mtl-and-ml-lifecyle}.

\section{Additional Multi-Task Learning Approaches}
\label{section:appendix-mtl-approaches}

\subsection{Prompts}
\label{subsection:appendix-prompts}

Prompts embed a task in the input. 
The original input $\mathbf{x}$ is modified using a template into a textual string prompt $\mathbf{x'}$ that has some unfilled slots, and then the LM is used to fill the information to obtain a final string $\mathbf{\hat{x}}$, from which the final output $\mathbf{y}$ can be derived \citep{liu2021pretrain}.
Prompting requires redesigning all the inputs and outputs in order to treat the tasks as text-to-text problems. 
The prompting approach proved to work the best in the zero- and few-show scenarios, but the benefits dissipate in the high-resource settings \citep{parmar2022inboxbart, wang2022instructionner}. Also, performance scales well with an increase in model size \citep{lester2021power}, making this approach not accessible to everyone.

According to \citet{liu2021pretrain}, there are many flavours to prompting. First, models with different pre-training objectives can be used -- left-to-right LM \citep{brown2020language}, MLM \citep{liu2019roberta}, prefix LM \citep{pmlr_v119_bao20a}, or encoder-decoder one \citep{lewis2020bart}. 
Prompts can be engineered in a cloze \citep{petroni2019language} or prefix \citep{lester2021power} shape, be hand-crafted \citep{brown2020language} or automated, which can again be discrete \citep{shin2020autoprompt} or continuous \citep{lester2021power}. 
Answer engineering searches for an answer space and a mapping to the original output by deciding the answer shape and choosing an answer design method. 
Furthermore, parameters can be updated using different settings -- tuning-free prompting \citep{brown2020language}, fixed-LM prompt tuning \citep{li2021prefix}, fixed-prompt LM tuning \citep{raffel2020exploring}, and prompt+LM tuning \citep{liu2021gpt}.
Finally, training can be done in a few/zero-shot \citep{brown2020language} or full-data setting \citep{han2021ptr}. In the following paragraphs, we discuss some of the previous prompting works. 

HyperPrompt \citep{he2022hyperprompt} is an approach that combines hypernetworks and prompts. The key idea is to prepend task-conditioned trainable vectors to both keys and values of MHA sub-layer at every Transformer layer. These task-specific attention feature maps are jointly trained with the task-agnostic representations. Key prompts interact with the original query, enabling tokens to acquire task-specific semantics. Value prompts are prepended to the original value vector, serving as task-specific memories for MHA to retrieve information from. However, instead of having a key/value prompt for each task and layer, authors initialize a global prompt $P$ for each task. They apply two local hypernetworks $h_{k/v}$ (one for keys, the other for values) at each Transformer layer in order to project this prompt into actual task- and layer-specific key/value prompts. There are three variations examined in the paper -- HyperPrompt-Share, -Sep, and -Global. HyperPrompt-Global proved to be the best, as it allows a flexible way to share information across tasks and layers. It introduces task and layer embeddings, which are then fused into the layer-aware task embedding. This embedding is then an input for the global hypernetworks $H_{k/v}$ used as a generator of local hypernetworks $h_{k/v}$. These local hypernetworks then generate hyper-prompts using a global prompt $P$. Hyper-prompts are finally prepended with the original keys and values in the MHA sub-layers. During training, no parameters are kept frozen. They report better results compared to the competitive methods of HyperFormer++ \citep{mahabadi2021parameter} and Prompt-Tuning \citep{lester2021power}.

In-BoXBART \citep{parmar2022inboxbart} uses biomedical instructions which contain: (1) definition (core explanation and detailed instructions about what needs to be done), (2) prompt (short explanation of the task), and (3) examples (input/output pairs with the explanation). Each instruction (effectively a prompt) is prepended to the input instances. A problem of too long instances arises.

InstructionNER \citep{wang2022instructionner} enriches the inputs with task-specific instructions and answer options. Additionally, two auxiliary tasks are introduced -- \textit{entity extraction} and \textit{entity typing}, which directly help in solving a NER task.

\citet{sanh2021multitask} allowed public to suggest prompts and came up with 2073 prompts for 177 datasets in total (12 prompts per dataset on avarage). They shuffled and combined all the examples from the datasets prior to training. In most of the cases, performance of their models improved as the number of training datasets increased. Moreover, training on more prompts per dataset resulted in a better and more robust generalization on unseen datasets. The models they trained are based on LM-adapted T5 model \citep{lester2021power}.

Prefix tuning freezes the language model parameters and optimizes a small, continuous task-specific vector called \textit{prefix} \citep{li2021prefix}. Consequently, only a prefix needs to be stored for each task, making prefix tuning modular and space-efficient. This approach can be applied solely to text generation models, such as GPT-2 \citep{radford2019gpt2} and BART \citep{lewis2020bart}. They state the intuition that the context can influence the encoding of input $\mathbf{x}$ by guiding what to extract from $\mathbf{x}$; and can influence the generation of output $\mathbf{y}$ by steering the next token distribution. They optimize the prefix as continuous word embeddings, instead of optimizing over discrete tokens. The reason for this is that the discrete prompt needs to match the real word embedding, resulting in a less expressive model. 

\citet{lester2021power} used a fixed-length prompt of special tokens, where the embeddings of these tokens are updated. This removes the requirement that the prompts are parametrized by the model and allows them to have their own trainable parameters. They test three different initialization techniques -- random uniform, sampled vocabulary, and class label. LM-adaptation pre-training technique was used. Unlike adapters, which modify the actual function that acts on the input, prompt tuning adds new input representations that affect how input is processed, leaving the function fixed. They freeze the pre-trained model. Finally, they try prompt ensembling (one prompt per model, for each task), which showed improved performance compared to a single-prompt average.

\textbf{Challenges and Opportunities in ML Lifecycle.} Prompts can also mitigate some of the challenges of different ML lifecycle phases (see Table~\ref{table:ml-lifecycle-phase-challenges}). 
To start with, in a fixed-prompt LM tuning or a prompt+LM tuning setting, knowledge is transferred from other tasks, which can be especially beneficial for low-resource tasks. However, in such a scenario, different optimization techniques need to be considered in order to avoid negative interference and other MTL drawbacks (Appendix~\ref{subsection:appendix-mtl-surveys}). As for the model selection, it is more of a challenge than an opportunity, as models based on prompts require large number of parameters to begin with in order to perform well. Model training complexity depends on the parameter update setting, with some settings requiring no \citep{brown2020language} or only few parameter updates \citep{li2021prefix}. One of the benefits when using prompts is the ability to handle all tasks using a single text-to-text model, regardless of the input and output encoding.
When using prompts, the biggest challenge in model deployment is the model's size. Finally, model updating due to distribution shift, new data, or business requirements (see Section~\ref{subsection:phase-model-update-problem}) seems most plausible in the setting where prompts are continuous and tuned \citep{li2021prefix}. However, this has downsides, such as difficult optimization, non-monotonic performance change with regard to the number of parameters, and reserving a part of sequence length for adaptation \citep{hu2021lora}.

\end{document}